\documentclass[10pt,twocolumn,letterpaper]{article}

\usepackage{cvpr}              %

\usepackage[dvipsnames]{xcolor}

\definecolor{cvprblue}{rgb}{0.21,0.49,0.74}
\usepackage[pagebackref,breaklinks,colorlinks,citecolor=cvprblue]{hyperref}
\usepackage{subcaption}
\usepackage{url}            %
\usepackage{booktabs}       %
\usepackage{amsfonts}       %
\usepackage{nicefrac}       %
\usepackage{microtype}      %
\usepackage{xcolor}         %
\usepackage{graphicx}
\usepackage{amsmath}
\usepackage{amsthm, amssymb}
\usepackage{float}
\usepackage{adjustbox}
\usepackage{placeins}
\newcommand{\pdfgrid}[4]{%
  \begin{adjustbox}{width=\textwidth}
    \begin{tabular}{@{}cc@{}}
      \includegraphics[width=0.5\textwidth]{qual/#1.pdf} &
      \includegraphics[width=0.5\textwidth]{qual/#2.pdf} \\
      \includegraphics[width=0.5\textwidth]{qual/#3.pdf} &
      \includegraphics[width=0.5\textwidth]{qual/#4.pdf} \\
    \end{tabular}
  \end{adjustbox}
}

\def\paperID{213} %
\def\confName{3DV\xspace}
\def\confYear{2026\xspace}

\title{Improved Convex Decomposition with Ensembling and Negative Primitives}

\author{
\begin{tabular}{c}
Vaibhav Vavilala$^{1}$ \quad
Florian Kluger$^{2}$ \quad
Seemandhar Jain$^{1,3}$ \quad
Bodo Rosenhahn$^{2}$ \\
Anand Bhattad$^{4}$ \quad
David Forsyth$^{1}$ \\
\\
\begin{tabular}{@{}c c@{}}
$^{1}$University of Illinois Urbana-Champaign &
$^{2}$Leibniz Universität Hannover \\
$^{3}$University of California, San Diego &
{$^{4}$Johns Hopkins University}
\end{tabular}
\end{tabular}
}

\begin{document}

\maketitle

\begin{abstract}
  Describing a scene in terms of primitives -- geometrically simple shapes that offer a parsimonious but accurate
  abstraction of structure -- is an established and difficult fitting problem. Different scenes require different
  numbers of primitives, and these primitives interact strongly.  Existing methods are evaluated by comparing predicted depth, normals, and segmentation against ground truth. The state of the art method involves a learned regression procedure to predict a start point
  consisting of a fixed number of primitives, followed by a descent method to refine the geometry and remove redundant
  primitives.
  
  CSG (Constructive Solid Geometry) representations are significantly enhanced by a set-differencing operation.  Our representation incorporates {\em
    negative} primitives, which are differenced from the positive primitives.  These notably enrich the geometry that
  the model can encode, while complicating the fitting problem.    This paper  presents a method that can
  (a) incorporate these negative primitives  and (b)  choose the overall number of positive and negative primitives by
  ensembling.   Extensive experiments on the standard NYUv2 dataset
  confirm that (a) this approach results in substantial improvements in depth representation and segmentation over SOTA
  and (b) negative primitives improve fitting accuracy.
  Our method is robustly applicable across datasets: in a first, we evaluate primitive prediction for LAION images.
\end{abstract}

\begin{figure}[!ht]
  \centering
  \includegraphics[width=1\linewidth]{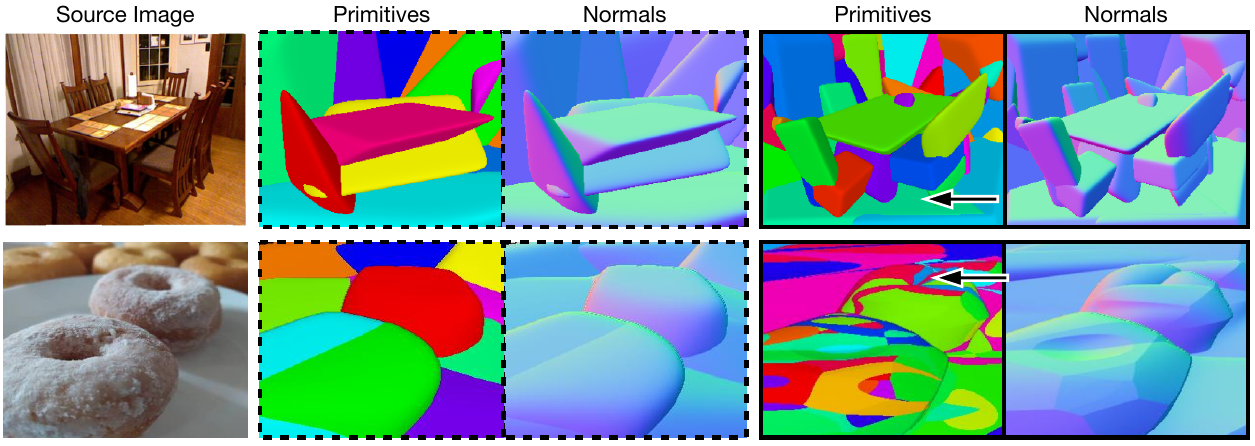}
  \caption{We present a method that can extract 3D primitives from RGB images with high geometric accuracy. Our method exploits negative primitives, which carve away geometry (i.e. a boolean operation). These subtractive operations enrich the shapes we can encode, such as holes in the donuts (\textbf{bottom row}). While no GT labels exist for the primitive parameters, we show how test-time ensembling can be used to find the appropriate level of abstraction for a given image. The second pair of results from each row can be obtained via ensembling, which produces more accurate geometric fits. Our test-time ensembling involves running several independent models and selecting the best solution for that image.}
  \label{fig:teaser}
\end{figure}

\section{Introduction}
\label{sec:intro}
Compact yet accurate representations of scene and object geometry have a wide range of applications,
including image editing, motion planning, and
grasping.  In image editing, a compact representation makes it easy for a user to select what
must be edited (e.g.~\cite{bhat2023loosecontrol,vavilala2025generativeblocksworldmoving}), and accuracy is required for tasks like camera movement (Figure~\ref{fig:b2w_2}).
In motion planning and grasping, compactness supports efficient computation and accuracy is required to produce solutions that work (e.g.~\cite{amatoprim,xu2025rgbsqgraspinferringlocalsuperquadric,sqgrasp,primgrasp,primgrasp2,tracy2023differentiablecollisiondetectionset}.
This paper describes a method that parses a scene into a set of simple, canonical shapes, or primitives.
Our method balances three criteria: (i) \emph{compactness} -- the scene is represented with few
primitives; (ii) \emph{accuracy} -- the geometric error between the representation and the ground truth scene is small;
and (iii) \emph{semantic coherence} -- primitive boundaries are approximately aligned with object or part boundaries.

We follow \cite{deng2020cvxnet} in using smoothed 3D polytopes, which avoid the geometric difficulties (e.g. singularities, non-convexities, self-intersections and symmetries)
associated with superquadrics.  Existing approaches to fit primitives are either descent methods (which optimize primitive parameters directly,
and so susceptible to poor initialization and local minima) or regression methods (which predict parameters in a
single pass, so can avoid the worst local minima but lack the precision to produce a tight fit to the specific input geometry).
In existing methods, model selection  (choosing the number of primitives for a scene) is by a single hyperparameter. But some scenes are
more complicated than others (\cref{fig:teaser} top row).  Further, existing methods lack the expressiveness to model concavities or holes efficiently (\cref{fig:teaser} bottom row).  In contrast to~\cite{deng2020cvxnet}, we use a hybrid regress then descent method; fit entire scenes using RGB input on a very large scale;
fit negative primitives, allowing set differencing to make concavities;  choose geometric complexity of the fitted representation
with an adaptive method built on ensembling; and significantly exceed SOTA accuracy with very small representations.
Our contributions are:
\begin{itemize}
    \item \textbf{Fitting CSG Models with Negative Primitives}: We present the first method capable of fitting CSG models that include a set differencing operator to in-the-wild images of scenes. We demonstrate significant quantitative and qualitative benefits to using negative primitives for scene decomposition.
    \item \textbf{Test-Time Ensembling for Model Selection}: We introduce a test-time search procedure that dynamically selects the number of positive and negative primitives for each scene. This data-driven approach yields a more accurate representation than any fixed-count model.
    \item \textbf{State-of-the-Art Accuracy:} Our combined method significantly outperforms all available baselines and prior state-of-the-art on the NYUv2 benchmark, achieving a relative error reduction of over 50\% and demonstrating strong generalization to diverse, in-the-wild scenes.
    \item \textbf{Viability at large scale:} We demonstrate fitting to over 1M in-the-wild LAION images with high accuracy.
\end{itemize}

\section{Related Work}
\label{sec:related}

Primitives date to the origins of computer vision.
Roberts worked with blocks~\cite{roberts}; Binford with
generalized cylinders~\cite{binford71};  Biederman with
geons~\cite{biederman}.  Ideally, complex
objects might be handled with simple primitives~\cite{Chen2019BSPNetGC} where
each primitive is a semantic part~\cite{biederman,binford71,lego}. 
Primitives can be recovered from image data~\cite{nevatia77,shgc}, and allow
simplified geometric reasoning~\cite{hebertponce}.

\begin{figure}[t]
    \centering
    \begin{subfigure}{0.48\textwidth}
        \includegraphics[width=\linewidth]{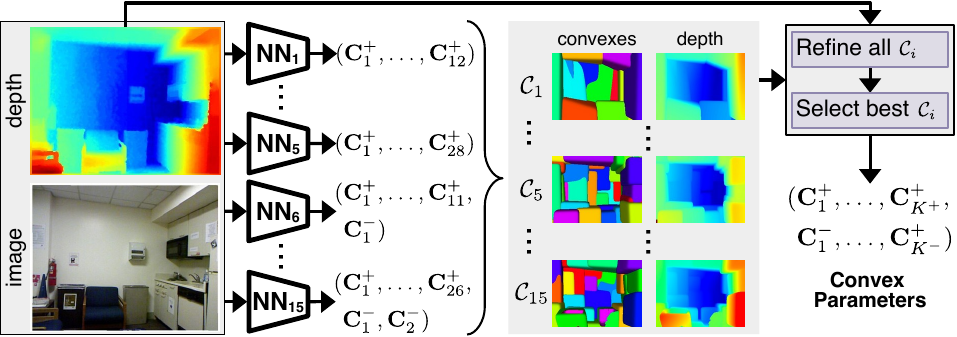}
        \caption{\textbf{Inference Overview:} An RGBD image is input to an ensemble of independently trained CNNs. Each network predicts the parameters of a set of convexes $\mathcal{C}_i$. The number of convexes varies between $12$ and $36$ in this work, with many of them potentially being negative. We refine each set of convexes by minimizing the training loss w.r.t. the input depth map. Our final decomposition consists of the set of refined convexes $\mathcal{C}_i$ which yields the lowest absolute relative depth error.}
        \label{fig:ensembling_pipeline}
    \end{subfigure}
    \hfill
    \begin{subfigure}{0.43\textwidth}
        \includegraphics[width=\linewidth]{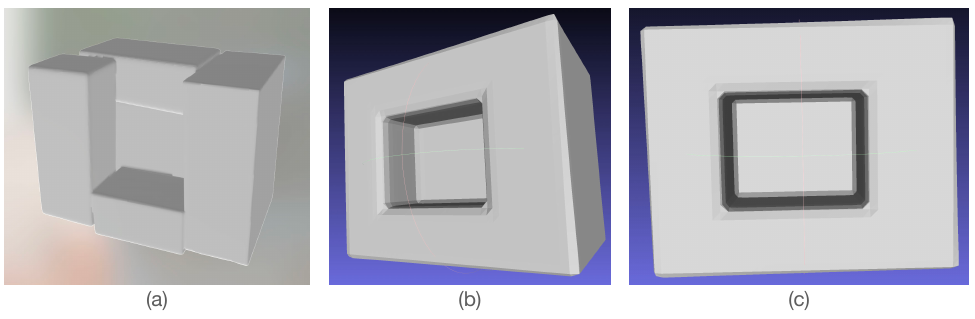}
        \caption{\textbf{Negative primitives are parameter-efficient.} Representing a simple box with a hole punched in it can be challenging even with several traditional primitives, as shown in \textbf{(a)}, where five primitives get stuck in a local minimum. In contrast, two primitives - one positive and one negative - can represent the geometry successfully because of the enriched vocabulary of operations. Two views are shown in \textbf{(b)} and \textbf{(c)}.}
        \label{fig:neg_intro}
    \end{subfigure}
    \caption{Overview of our approach (left) and demonstration of negative primitives (right).}
    \label{fig:combined}
\end{figure}

For individual objects, neural methods could predict the right set of primitives 
by predicting solutions for test data that are ``like'' those that worked for training data. Tulsiani {\em et
al.}  parse 3D shapes into cuboids, trained without ground truth segmentations~\cite{abstractionTulsiani17}. Zou {\em et al.} parse with a recurrent architecture \cite{Zou_2018_CVPR}. Liu {\em et al.} produce detailed reconstructions of objects in indoor scenes, but do not attempt parsimonious
abstraction~\cite{liu2022towards}. Worryingly, 3D
reconstruction networks might rely on object
semantics~\cite{Tatarchenko2019WhatDS}. Deng {\em et al.} (CVXNet)
represent objects as a union of convexes, again training without
ground truth segmentations~\cite{deng2020cvxnet}. An early variant of CVXNet 
can recover 3D representations of poses from single images, with reasonable parses into parts~\cite{deng2019cerberus}.
Meshes can be decomposed into near convex primitives, by a form of search~\cite{weiacd}.
Part decompositions have attractive editability~\cite{spaghetti}. Regression methods face some difficulty producing
different numbers of primitives per scene (CVXNet uses a fixed number;~\cite{abstractionTulsiani17} predicts the
probability a primitive is present; one also might use Gumbel softmax~\cite{jang2017categorical}).
Primitives that have been explored include: cuboids~\cite{Calderon2017BoundingPF,Gadelha2020LearningGM,Mo2019StructureNetHG,abstractionTulsiani17,Roberts2021LSDStructureNetML,Smirnov2019DeepPS,Sun2019LearningAH,kluger2021cuboids};
superquadrics~\cite{Barr1981SuperquadricsAA,Jakli2000SegmentationAR,Paschalidou2019SuperquadricsRL}; planes~\cite{Chen2019BSPNetGC,Liu2018PlaneRCNN3P}; and generalized cylinders~\cite{nevatia77, Zou20173DPRNNGS, Li2018SupervisedFO}.
There is a recent review in~\cite{fureview}.

Neural Parts~\cite{paschalidou2021neural} decomposes an object given
by an image into a set of non-convex shapes.
CAPRI-Net~\cite{yu2022capri} decomposes 3D objects given as point
clouds or voxel grids into assemblies of quadric surfaces.
DeepCAD~\cite{wu2021deepcad} decomposes an object into a sequence of
commands describing a CAD model, but requires appropriately
annotated data for training. Point2Cyl~\cite{uy2022point2cyl} is similar, but predicts the 2D shapes as an SDF.
Notably, \cite{yu2022capri,wu2021deepcad,uy2022point2cyl} also utilize CSG with negative parts but, unlike our work, focus on CAD models of single objects instead of complex real-world scenes.

Hoiem {\em et al.} parse outdoor scenes into vertical and horizontal
surfaces~\cite{hoiem:2005:popup,hoiemijcv2007}; Gupta {\em et al.}
demonstrate a parse into blocks~\cite{s:gupta10}. Indoor scenes can be
parsed into: a cuboid~\cite{hedauiccv2009,Vavilala_2023_ICCV}; beds and some furniture as
boxes~\cite{hedaueccv2010}; free space~\cite{hedaucvpr2012}; and plane
layouts~\cite{stekovic2020general, liu2018planenet}. If RGBD is
available, one can recover layout in detail~\cite{Zou_2017_ICCV}. Patch-like primitives can be imputed from
data~\cite{Fouhey13}. Jiang demonstrates parsing  RGBD images into
primitives by solving a 0-1 quadratic
program~\cite{jiang2014finding}. Like that work, we evaluate
segmentation by primitives (see~\cite{jiang2014finding}, p. 12), but
we use original NYUv2 labels instead of the drastically simplified
ones in the prior work. Also, our primitives are truly convex. 
Monnier {\em et al.} and Alaniz {\em et al.} decompose scenes into sets of superquadrics using differentiable rendering, which requires calibrated multi-view images as input~\cite{monnier2023differentiable,alaniz2023iterative}.
Most similar to our work is that of Kluger {\em et al.}, who identify
cuboids sequentially with a RANSAC-like
algorithm~\cite{ransac,kluger2020consac,kluger2021cuboids, kluger2024robust,kluger2024parsac}. 

The success of a descent method depends critically on the start point, typically dealt with using
greedy algorithms (rooted in RANSAC~\cite{ransac}; note the prevalence of RANSAC in a recent review~\cite{Kang2020ARO});
randomized search~\cite{Ramamonjisoa2022MonteBoxFinderDA,Hampali2021MonteCS}; or multiple starts.  Remarkably, as
Sec.~\ref{polishing} shows, modern first-order methods are capable of producing fairly good primitive
representations from a random start point~\cite{loshchilov2019decoupledweightdecayregularization}.
 
\begin{figure*}[t]
    \centering
    \includegraphics[width=\linewidth]{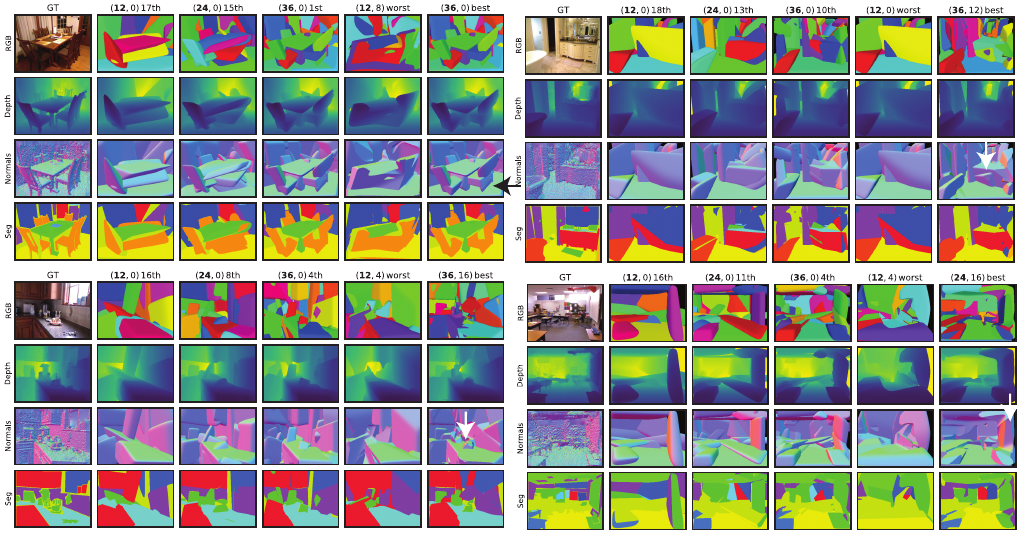}%
    \caption{Visualizations of various primitive predictions for four scenes from NYUv2.  We show
    ground truth (first column in each block); predictions of $(\mathbf{12}, \mathit{0})$, $(\mathbf{24}, \mathit{0})$ and $(\mathbf{36}, \mathit{0})$ models; the prediction
    of the model that made the \textit{worst} prediction for the scene; and the prediction of the model that made the \textit{best}
    prediction.  The best choice of primitive counts $K^{total}$ varies from scene to scene.
    Notice some complex objects made up as composites of positive primitives (black arrow) and negative primitives
    ``carving out'' shapes.  The segmentation label is the oracle label described in the text.
    }
    \label{fig:qualNYUv2}
\end{figure*}

\begin{figure*}
    \centering
    \pdfgrid{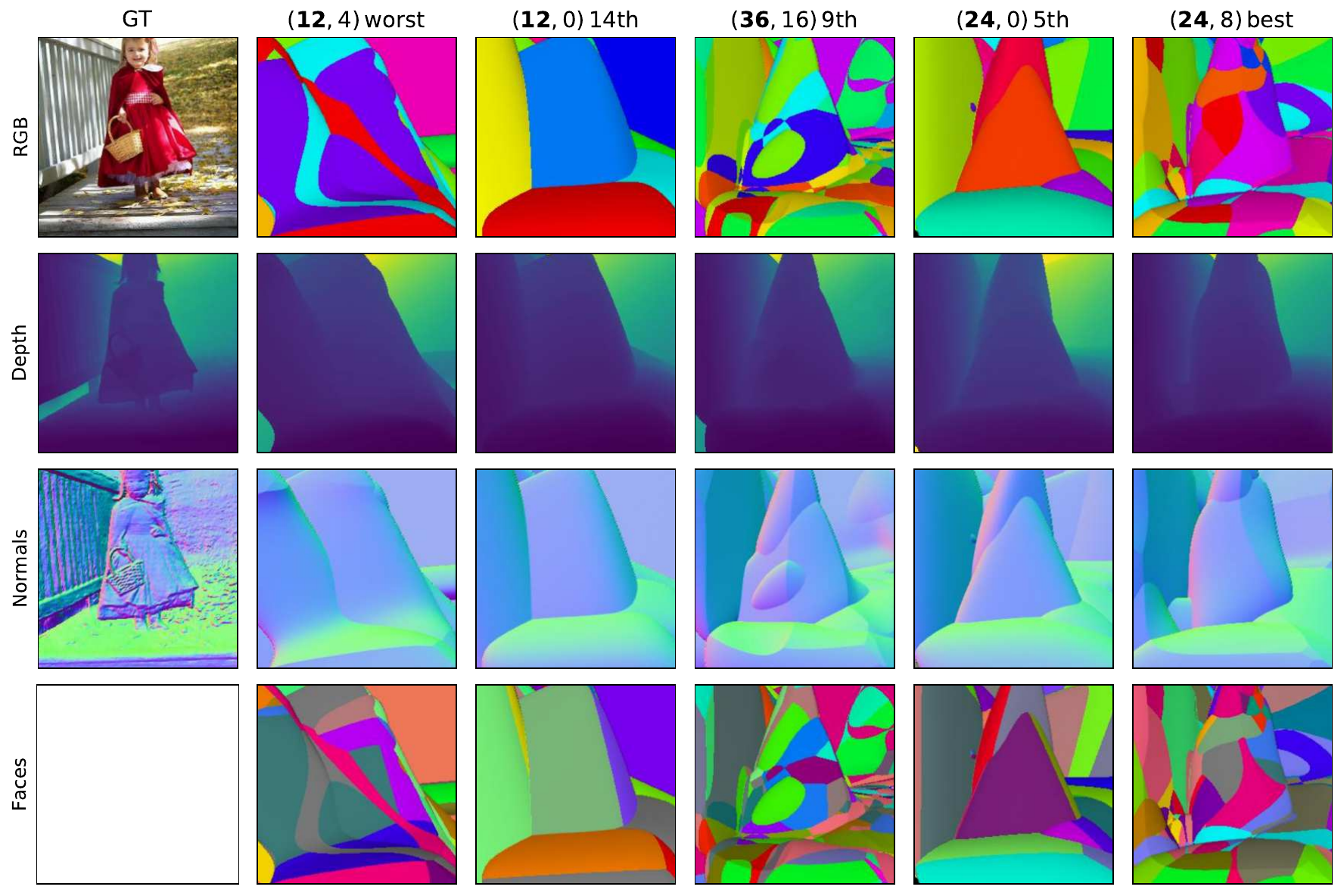}{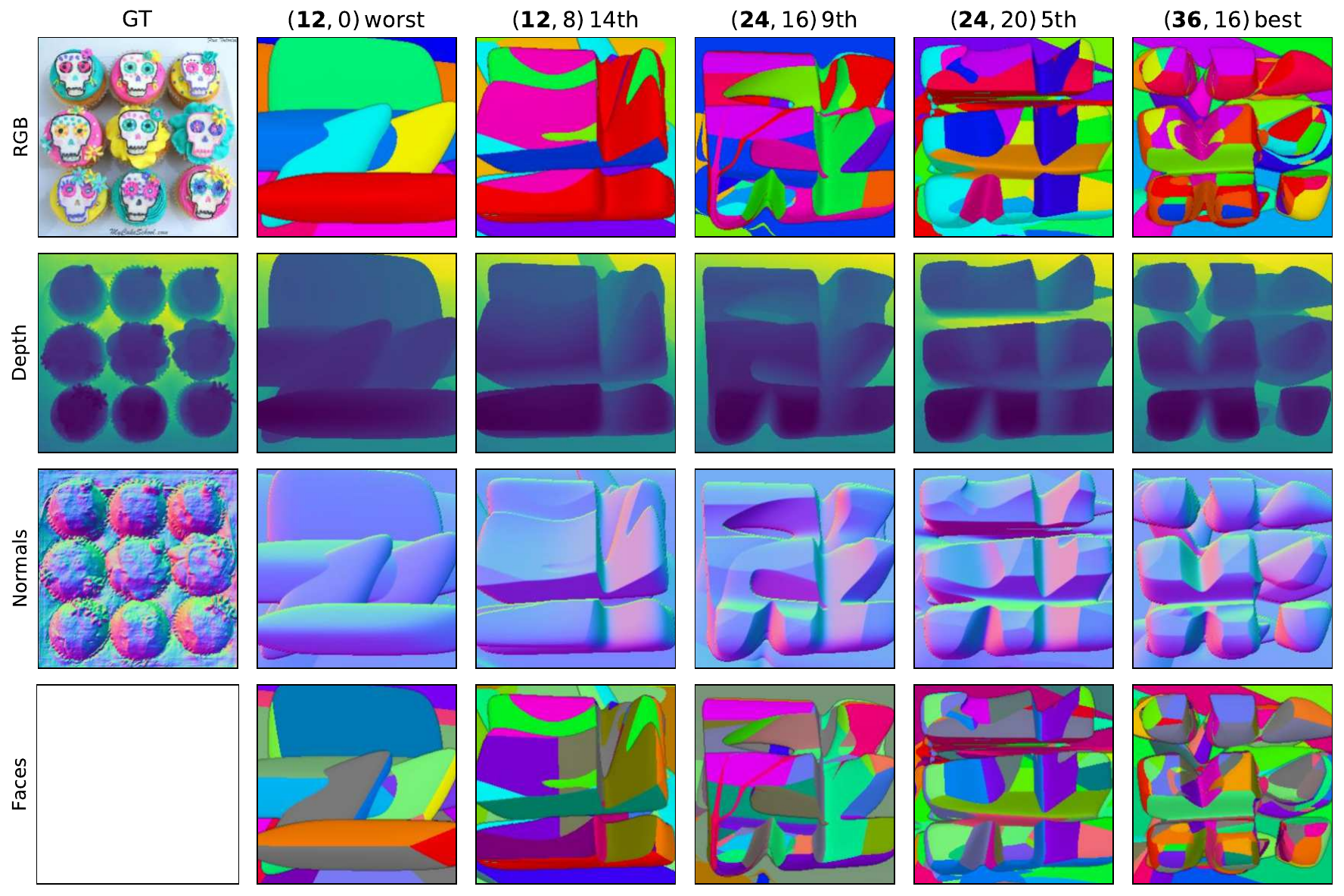}{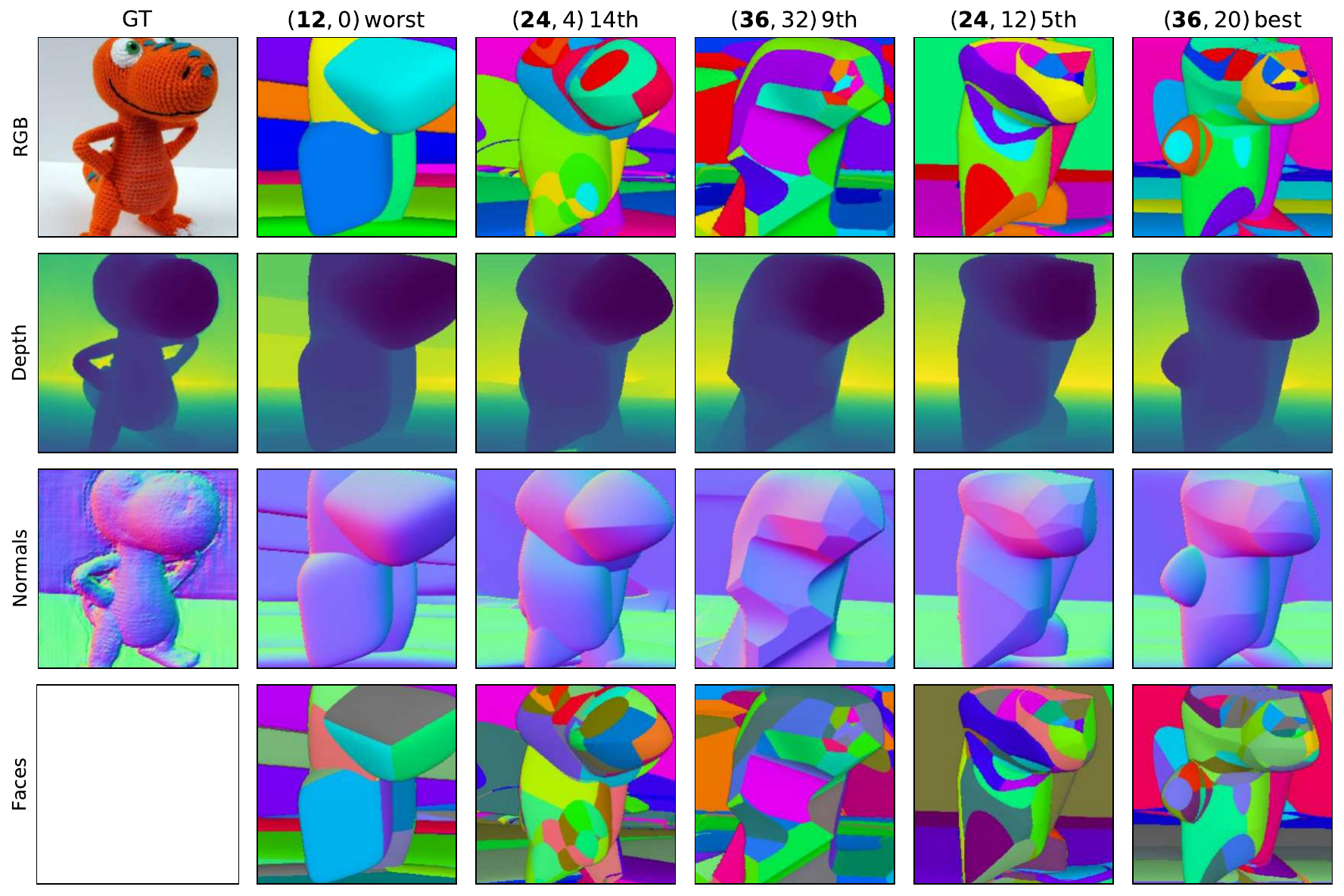}{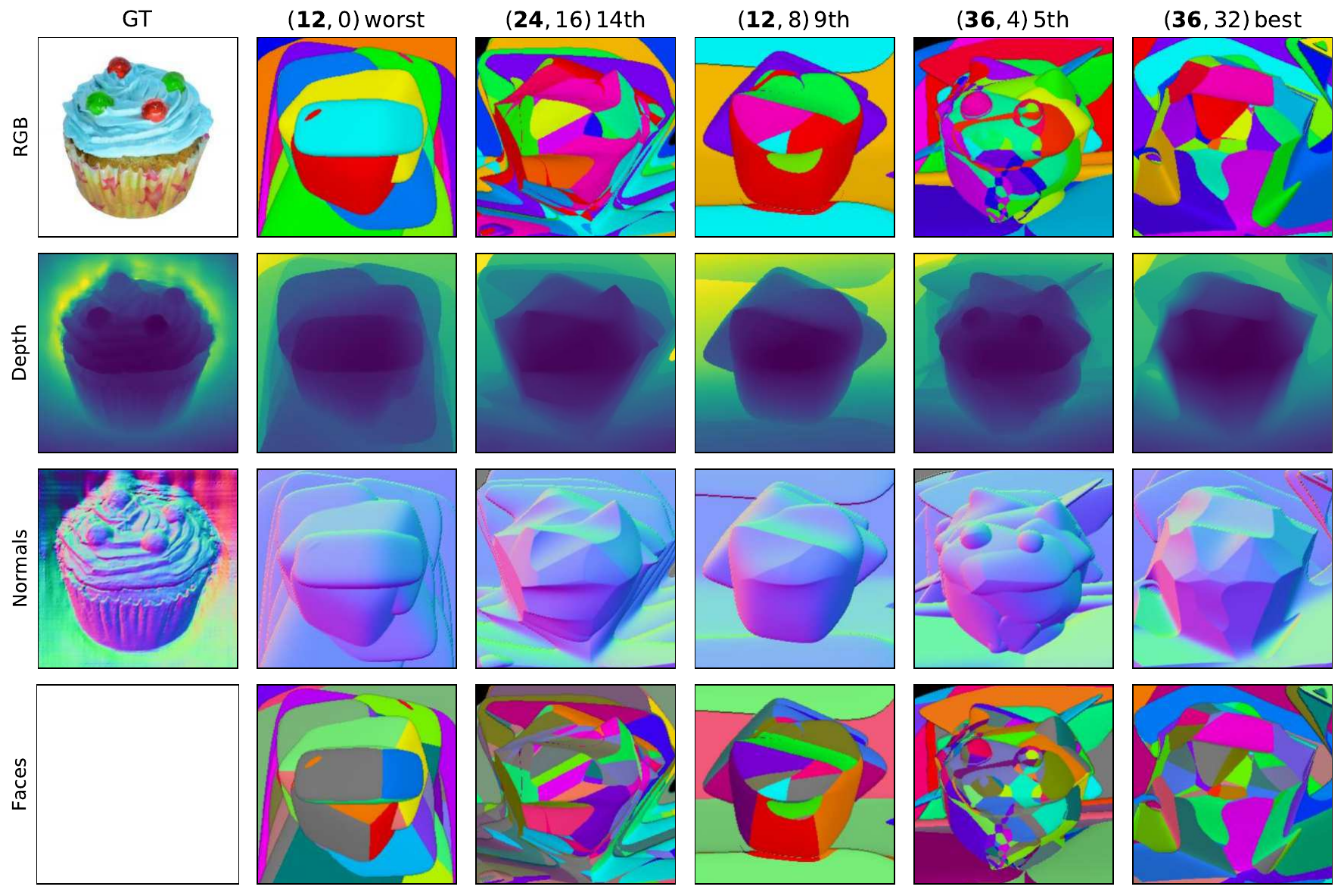} 
    \caption{Visualizations of various primitive predictions for four scenes from LAION.  We show
    ground truth (first column in each block); the prediction
    of the model that made the \textit{worst} prediction for the scene; the prediction of the model that made the \textit{best}
    prediction rightmost; and in-between results shown in intermediate columns. While our models produce good fits on average, ensembling helps select the optimal primitive count for each test scene and avoids poor solutions. The best choice of primitive counts $K^{total}$ varies from scene to scene.  {\bf Bottom row} in each block shows face labels -- no oracle segmentation is available.
    Note how primitives can follow complex structures; how they tend to ``stick'' to object properties; and how the number of face labels grows very quickly with
    the number of negative primitives. See Figs.~\ref{fig:quallaion2} and~\ref{fig:quallaion3} for additional examples.}
    \label{fig:quallaion}
\end{figure*}

\begin{figure*}[t!]

\centering
\includegraphics[width=0.85\textwidth]{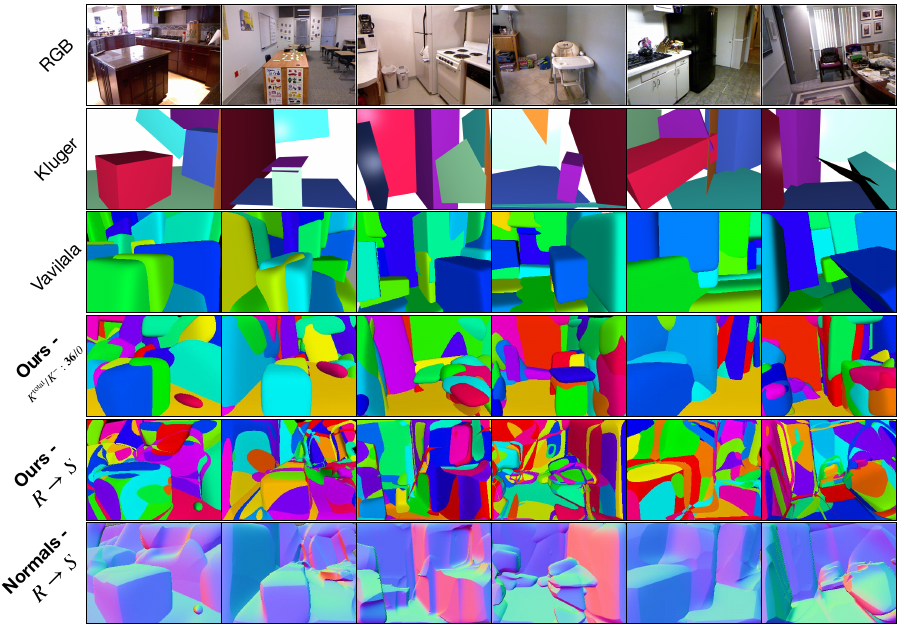}
\caption{We present a method that advances the SOTA for primitive decomposition by using ensembling and negative primitives. Prior art shown in $\mathbf{2^{nd}}$ row~\cite{kluger2021cuboids} and $\mathbf{3^{rd}}$ row~\cite{Vavilala_2023_ICCV} for comparison. (\textbf{4th row}) We show results from our approach with $\mathbf{36}$ primitives, none negative. Our procedure encodes geometry quite closely. (\textbf{5th and 6th row}) With negative primitives, we can encode a rich arrangement of shapes. Here, we ensemble many predictors and show the best one. In these six cases, a model with negative primitives was chosen. The normals make it clear that negative primitives are scooping geometry away from positive primitives.}
\label{fig:teas}
\end{figure*}

\section{Method}
\label{sec:method}
\label{sec:boolean_primitives}
 Write $K^{total}$ for the total number of primitives and $K^{-}$ for the number of negatives to be predicted.
 For each $(K^{total}, K^{-})$ we wish to investigate, we train a prediction network.   Then, at inference time, we produce a
set of primitives from each network, polish it,  select the primitives with the best loss and report that, so
different scenes will have predictions involving different numbers of primitives. Our approach generalizes the architecture of~\cite{Vavilala_2023_ICCV}, but produces very significant improvements in
 performance (Sec.~\ref{sec:results}). Negative primitives require  some minor modifications of their procedure (Sec.~\ref{sec:boolean_primitives}).
 Our losses (Sec.~\ref{sec:losses}) and test-time refinement (Sec.~\ref{polishing}) are somewhat similar.

 Our network requires RGBD input. If depth measurements are not available, our method works well with single image depth predictors~\cite{Ranftl2022, depthanything}. Our losses require a point cloud that is extracted from the depth image via the heuristic
 described in ~\cite{Vavilala_2023_ICCV}. Fig.~\ref{fig:ensembling_pipeline} provides an overview of our inference pipeline.

 {\bf Base primitives:} Our primitives are smoothed polytopes as described in~\cite{deng2020cvxnet}. For 6-faced
 parallelepipeds, each primitive is parametrized by a center (3 DOF's), 3 normals (6 DOF's), 6 offsets (6 DOFs) and a
 blending term (1 DOF). The blended half-plane approach eases training and also enables fitting curved surfaces. We fit 6-faced parallelepipeds for fair comparative evaluation, and then show that more faces per polytope yields better representations (see Table~\ref{tab:auc_laion_12} in  supplementary).

 {\bf Negative primitives:}  Set differencing produces a notably more complex geometric representation.  Assume we have
 $K^{total}$ primitives of which $K^-$ are negative, each with $f$ faces.  Label an image pixel by the face intersection
 that produced that pixel (as in our face segmentation figures, e.g. Fig.~\ref{fig:quallaion}). Generic pixels could result
 from either ray intersection with a face of a positive primitive or with a face of a negative primitive inside some
 positive. This argument means that there are a maximum of $f\times (K^{total}-K^-) \times (1 +K^-)$ pixel labels; note how
 this number grows very quickly with an increase in the number of negative primitives, an effect that can be seen in Fig.~\ref{fig:quallaion}.
 Negative primitives are easily handled with indicator functions.
 We define the indicator for a set of  primitives  $O : \mathbb{R}^3 \rightarrow [0,1]$, with
$O(x) = 0$ indicating free space, and $O(x) = 1$ indicating a query point $x \in \mathbb{R}^3$ is inside the
volume. Write $O^+(x)$ for the indicator function for the set of positive primitives, $O^-(x)$ for negative
primitives. The indicator for our representation is then: 
\begin{equation}
O(x) = relu(O^+(x) - O^-(x)).
\end{equation}

As we will show in our results, Section~\ref{sec:results}, negative primitives allow us to encode a richer set of shapes, given the same primitive budget. Our analysis leads to the following theorem, which we analyze in the supplement Sec.~\ref{sec:theory}:
\newtheorem{theorem}{Theorem}
\begin{theorem}
  For the vocabulary of primitives described,  there exist shapes $S$ such that $K_{\pm}(S) \ll K_+(S)$ when the representation of $S$ is
  required to achieve sufficiently high precision.
\end{theorem}

\subsection{Losses}
\label{sec:losses}

Our modified representation allows re-using the existing sample loss and auxiliary losses (unique parametrization loss,
overlap loss, guidance loss, localization loss) ~\cite{deng2020cvxnet,Vavilala_2023_ICCV} for both $O^+(x)$ and
$O^-(x)$.   While a Manhattan World loss was found to be helpful for NYUv2, it hurt quality on general in-the-wild LAION images in
our testing so we exclude it. We do not consider the volume loss or segmentation loss
from~\cite{Vavilala_2023_ICCV} in our experimentation, as they were shown to have an approximately neutral effect in
the original paper.

\subsection{Polishing and Descent}
\label{sec:performance_improvements}
\label{polishing}

Test-time finetuning is possible because we can evaluate the primitive prediction against the predicted depth map,
then use the training losses at test-time.  The fit is improved by using more 3D samples in these losses per image at test-time.
Our polishing procedure is heavily optimized.

In fact, our test-time refinement procedure is effective enough that we can fit a primitive representation \textit{using only a random
  start}.   We are aware of no other primitive fitting procedure that can operate with pure descent and no random
restart or incremental process.  This supplies an interesting baseline; Sec.~\ref{sec:results} demonstrates that this
baseline is highly inefficient compared to polishing a network prediction, and is not competitive in accuracy.   

\subsection{Choosing the Number of Primitives}
\label{sec:ensembling}

Much of the literature on primitive decomposition fits a fixed number of primitives~\cite{deng2020cvxnet}.  In contrast,
we train 18 models for $(K^{total}, K^-)$, with $K^{total} \in \{\mathbf{12}, \mathbf{24}, \mathbf{36}\}$ and $K^- \in \{\mathit{0}, \mathit{4}, ..., \mathit{K}^{total}-\mathit{4}\}$. We investigate two strategies for choosing the best prediction (and so the best
set of primitives) for a given test image:  $S\rightarrow R$, where we select the best neural prediction then refine it;
and $R\rightarrow S$, where we refine all predictions then select the best.

\subsection{Implementation Details}
\label{sec:implementation_details}
Our neural architecture is a ResNet-18 encoder (first layer expanded to accept RGBD input), followed by a decoder consisting of three linear
layers of sizes $[1048, 1048, 2048]$ and LeakyRelu activations. We do not freeze any layers during training. The
dimensionality of the final output varies based on the number of primitives the model is trained to produce (as we train
different models for different numbers of primitives in this work). We implement our procedure in PyTorch and train all
networks with AdamW optimizer, learning rate $2\times10^{-4}$, batch size 96, mixed-precision training, for 20000
iterations, on a single A40 GPU. Each image is resized to $240\times320$ resolution. Although we train at fixed
resolution, our model can run inference at variable aspect ratio, as would be expected from CNNs like ResNet. It takes
39 mins to train a 12 primitive model and 62 mins to train a 36 primitive model. On LAION, we train at $256\times256$
resolution. We use random flip, scale, and crop during training. 

\begin{table*}[t]
\centering
\resizebox{\linewidth}{!}{%
\begin{tabular}{l c c c c c c c c}
\toprule
Method & $K^{total}$ & $K^-$ & AbsRel$\downarrow$ & Normals Mean$\downarrow$ &Normals Median$\downarrow$ & SegAcc$\uparrow$ & Time (s) & Mem (GB) \\
\midrule
12 & $\mathbf{12}$ & $\mathit{0}$ & 0.0622 & \textit{34.4} & 26.3 & 0.651 & 0.84 & 3.53 \\
24 & $\mathbf{24}$ & $\mathit{0}$ & 0.052 & 33 & 25 & 0.7 & 1.46 & 5.57 \\
36 & $\mathbf{36}$ & $\mathit{0}$ & 0.0484 & 32.3 & 24.4 & 0.723 & 2.06 & 7.61 \\
best & $\mathbf{36}$ & $\mathit{12}$ & \textit{0.0452} & \textit{32.1} & \textit{24} & \textbf{0.765} & 2.06 & 7.61 \\
\midrule
$\mathsf{pos}$ $S\rightarrow R$ & $\mathbf{26.1}$ & $\mathit{0}$ & 0.0527 & 33 & 24.9 & 0.7 & 2.08 & 7.61 \\
$\mathsf{pos+neg}$ $S\rightarrow R$ & $\mathbf{29}$ & $\mathit{10.4}$ & 0.0492 & 32.9 & 24.7 & 0.733 & 2.13 & 7.61 \\
$\mathsf{pos}$ $R\rightarrow S$ & $\mathbf{33.6}$ & $\mathit{0}$ & 0.0476 & 32.3 & 24.4 & 0.72 & 4.37 & 7.61 \\
$\mathsf{pos+neg}$ $R\rightarrow S$ & $\mathbf{35}$ & $\mathit{14.7}$ & \textbf{0.0417} & \textbf{31.9} & \textbf{24} & 0.756 & 29.9 & 7.61 \\
\midrule
Vavilala \& Forsyth~\cite{Vavilala_2023_ICCV} & $\mathbf{13.9}$ & $\mathit{0}$ & 0.098 & 37.4 & 32.4 & 0.618 & 40 & 6.77 \\
\bottomrule
\end{tabular}
}
\caption{Comparison to SOTA (last row) on NYUv2.  \textbf{All rows except the last were trained using our procedure.} Our best approach (second last row) polishes then chooses from 18
  different models with different numbers of primitives.
  \textbf{First three rows}: we train a primitive generation model according to the procedure laid out in
  Sec~\ref{sec:method}, without negative primitives. Next row: $\mathbf{36}$ total primitives with $\mathit{12}$ negative was our best network
  as measured by AbsRel.  \textbf{Next four rows} Ensembling strongly improves error metrics, particularly
  AbsRel. $\mathsf{pos+neg}$ refers to all 18 models available for ensembling, whereas $\mathsf{pos}$ refers to only 3 models without negative
  primitives available. $S \rightarrow R$ refers to only refining the output of the model with the best sample
  classification; $R \rightarrow S$ means we finetune all models and pick the best one. In this table, we finetune
  assuming GT depth is available at test time, though our method still works even when depth is inferred by a pretrained
  depth estimator. The fact that substantial gains can be achieved from $R\rightarrow S$ implies that the best start
  point may not yield the best end point -- meaning the fitting problem is hard. Time and memory estimates are presented
  as well. \textbf{Last row}: we compare our methods against existing work. Any individual model we train obtains better
  error metrics with less compute. Timings for ensembling show estimated total cost of running all the methods and
  selecting the best one; memory refers to peak GPU memory usage.} 
\label{tab:Seg_table}
\end{table*}

\begin{figure*}[t]
    \centering
    \includegraphics[width=\linewidth]{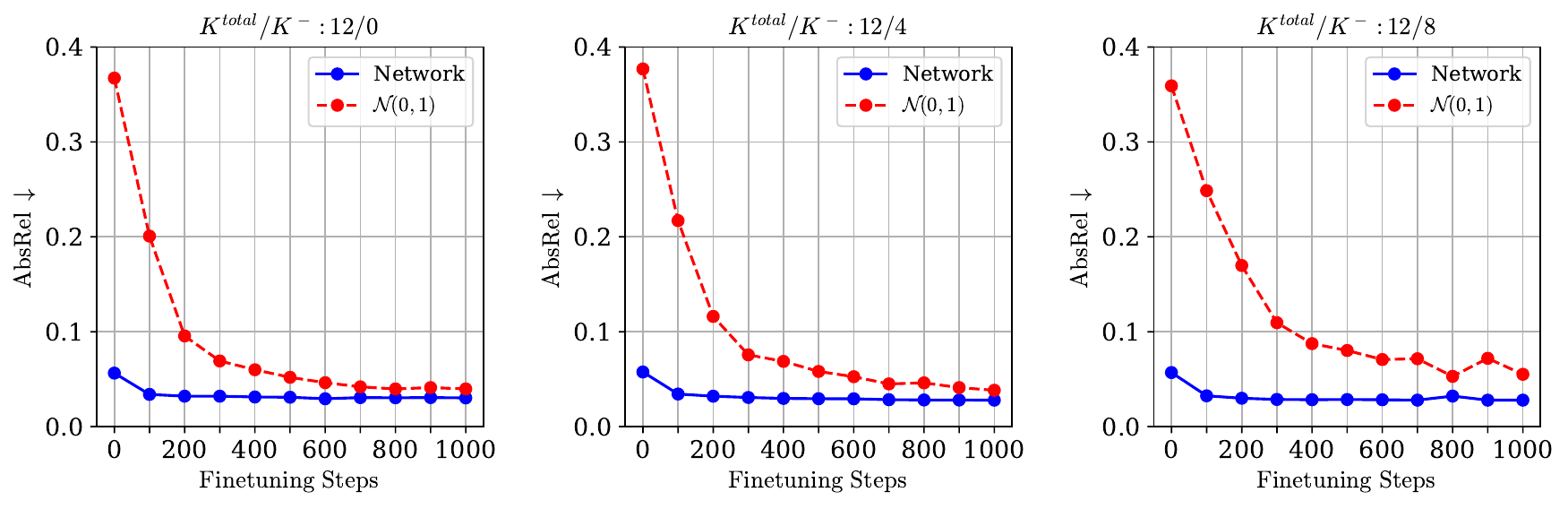}%
    \caption{\textbf{Network start is beneficial.} Initializing our finetuning process with primitives predicted by our
      network (blue line) yields better primitives after finetuning than random start (red line). In practice, network start saves around 900 FT steps and can achieve a better quality than random
      start. It also appears to be harder to fit negative primitives than positives (there is a larger gap in the final
      AbsRel between the two curves when there are negative primitives).  Based on these results, we use 200 finetuning
      steps to balance compute and quality when reporting error metrics in this paper. Further, the fact that it is even
      possible to generate high-quality primitives via pure optimization, without a neural network, is new in the
      context of recent primitive-generation literature. Results shown on 100 random test images from LAION.} 
    \label{fig:FineTune12}
\end{figure*}

\section{Results}
\label{sec:results}

Qualitative results appear in Fig~\ref{fig:qualNYUv2} and Fig~\ref{fig:quallaion}.
Note how primitives can combine to form complex structures; how negative primitives ``carve out'' complex shapes; how primitives tend to ``stick'' to object properties (for example, heads; a house); and how the number of face labels grows very quickly with the number of negative primitives.

\subsection{Evaluating a Primitive Representation}

While {\em producing} a primitive representation has a long history (\cite{MarrNishihara1978},
Sec.~\ref{sec:related}), not much is known about how one is to be {\em used} apart from the original recognition
argument, now clearly an anachronism.  Recent work in conditioned image synthesis (\cite{vavilala2025generativeblocksworldmoving,bhat2023loosecontrol}) suggests that applications might need
(a) a relatively compact representation (so that users can, say, move primitives around) and (b) one that accurately
reflects  depth, normals and (ideally) segmentation.

We compare primitive methods against one another using standard metrics for depth, normal and segmentation.
Specialized predictors of depth, normal and segmentation outperform primitive methods on these metrics.  But
we would not use a primitive predictor to actually predict depth, normal or segmentation -- instead, we are using
the metrics to determine whether very highly simplified representations achieve reasonable accuracy.
Our procedure uses the standard 795/654 train/test NYUv2 split~\cite{Silberman:ECCV12}. We hold out 5\% of training images for validation. We use this
 dataset primarily to maintain consistency in evaluating against prior art.  

For NYUv2, we compare the depth map predicted by primitives to ground truth using a variety of metrics; normals predicted
by primitives to ground truth; and a segmentation derived from primitives to ground truth segmentation. Depth
metrics are: the (standard) AbsRel (eg~\cite{Ranftl2020}); $AUC_n$, which evaluates the fraction of points within $n$ cm
of the correct location (after~\cite{Vavilala_2023_ICCV, kluger2021cuboids}); mean and median of the occlusion-aware
distance of~\cite{kluger2021cuboids}.  Normal metrics are based on~\cite{wang2015designing} and are mean and median of
angle to true normal, in degrees.  The segmentation metric uses the GT segmentation to assign the best label for each
image region, where regions consist of pixels with the same face intersection label (of
Sec.~\ref{sec:boolean_primitives}), then compares this to ground truth. The SegAcc error metric shown in Tables~\ref{tab:Seg_table} and~\ref{tab:nyu_full} refers to the fraction of pixels labeled correctly. For LAION, we compute depth and normal metrics from generated primitives comparing to depth and normal predicted from the image.

\begin{table}[htbp]

\centering

\begin{tabular}{c|c|c}
\toprule
Dataset& faces & ${\text{AbsRel}}{\downarrow}$ \\
\midrule

NYUv2 &6 & {0.0417}  \\
LAION& 6 & {0.0193}  \\
LAION& 12 & {0.0178}  \\

\bottomrule
\end{tabular}

\caption{Depth metrics for NYUv2 data and LAION data compared, when ensembling $ \mathsf{pos+neg} $ $R \rightarrow S$.
  The much larger data volume in LAION improves the generalization ability of our procedure. Further, increasing the representational power of our method by removing the symmetric normal constraint (for parallelepipeds, top two rows) and increasing the number of faces per polytope to $f=12$ (bottom row) yields even better primitive decompositions.
}
\label{tab:nyuv2vsLAION}
\end{table}

\subsection{NYUv2 Results}

{\bf Our method beats SOTA} on depth, normal, and segmentation (Table~\ref{tab:Seg_table}). With or without ensembling
process, {\bf our method is faster than SOTA}. Qualitative results in Fig.~\ref{fig:qualNYUv2}.  More extensive detailed comparison in Supplementary.

\begin{figure}[tbp]

\centering
\begin{subfigure}[t]{\linewidth}
    \centering
    \includegraphics[page=1, width=0.8\linewidth]{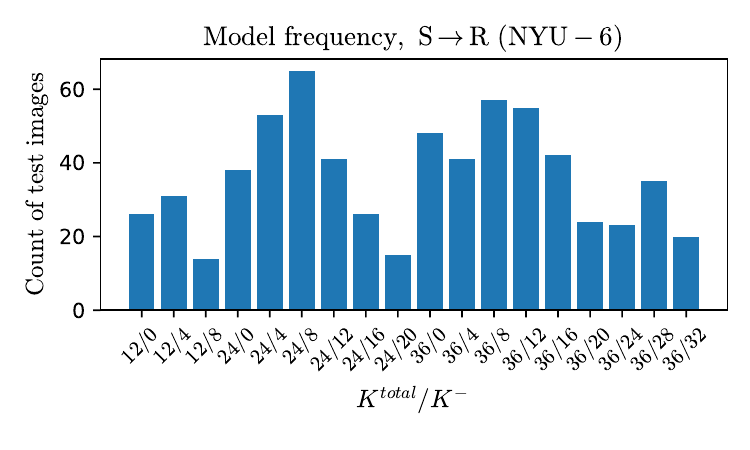}
    \vspace{-1em}
    \caption{Select then refine ensembling on NYUv2.}
    \label{fig:nyu_hist_a}
\end{subfigure}

\begin{subfigure}[t]{\linewidth}
    \centering
    \includegraphics[page=1, width=0.8\linewidth]{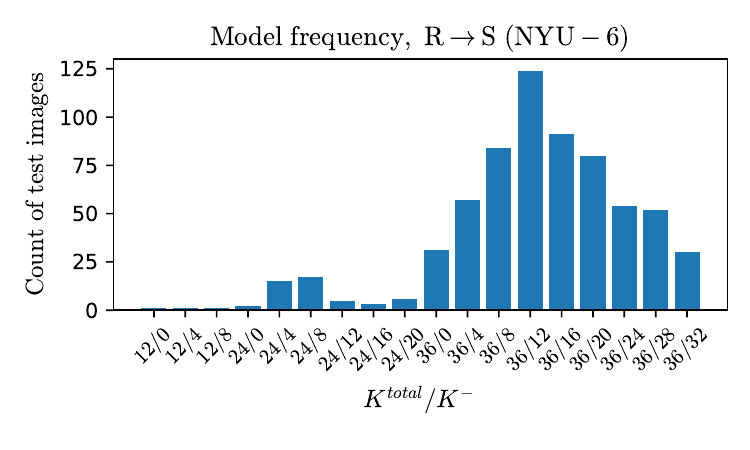}
     \vspace{-1em}
    \caption{Refine then select ensembling on NYUv2}
    \label{fig:nyu_hist_b}
\end{subfigure}
    \caption{Negative primitives are often selected from the ensemble. \textbf{Top} When we ensemble with $S\rightarrow R$, all models across all primitive counts are well-represented. This indicates that our network prediction may slightly struggle to manage larger numbers of primitives, hence the relative success of fewer primitives. In this setting, selecting a prediction for finetuning is based on fraction of 3D samples classified incorrectly, which is fast as we don't need to finetune and render the outputs of all the networks to decide which model to use. \textbf{Bottom} When we refine then choose $R \rightarrow S$, our finetuning procedure polishes each network start point and chooses the best one based on AbsRel, requiring a render for each model. When doing so, the best model (measured by AbsRel of rendered depth against GT depth) is strongly concentrated among higher primitive counts, $K^{total}=\mathbf{36}$.}
    \label{fig:nyu_hist}
\end{figure}
{\bf Negative primitives improve accuracy}, as indicated by Fig.~\ref{fig:nyu_hist_b}. This figure shows the
histogram of the number of times a particular $(K^{total}, K^-)$ combination was selected. Note that there is a strong
tendency to use more primitives ($K^{total}=\mathbf{36}$ is much more popular than others). One would expect this from bias
considerations, but we observe that many scenes use fewer primitives. The number of negatives used for the best fit is quite variable, though often near $K^- \approx \frac{1}{3}K^{total}$ - see Sec.~\ref{sec:theory}. We believe the ensembling step reduces variance by discarding complex fits that, while more flexible and thus lower in bias, tend to overfit and become less accurate. When complex fits perform poorly, the ensemble instead favors simpler, more robust models.

\begin{table}[h]
    \centering

    \begin{tabular}{c|c|c|c|c}
        \hline
        $K^{total}$ & Encode  & Loss & Finetune & Render  \\
        \midrule
        $\mathbf{12}$ & 0.0006 & 0.0006 & 0.68 & 0.15 \\
        $\mathbf{24}$ & 0.0006 & 0.0006 & 1.23 & 0.22 \\
        $\mathbf{36}$ & 0.0006 & 0.0007 & 1.79 & 0.26 \\
        \bottomrule
    \end{tabular}
    \caption{Estimated inference breakdown times, all times in seconds, 256-res images, $K^-=0$. Encoding is very fast, in which the network predicts parameters of the primitives given an RGBD image. Computing loss, required for getting the fraction of samples classified correctly when ensembling with $S \rightarrow R$, is also fast. However, finetuning (we show 200 steps here) is often the bottleneck since we must compute the loss and optimize the parameters of the primitives.}
    \label{tab:timing_breakdown}
\end{table}

{\bf Our method is efficient}, as Table~\ref{tab:timing_breakdown} shows.  The vast majority of time is spent in polishing the representation. Ensembling yields further improvements (particularly $S \rightarrow R$), and is still more efficient than prior work (see table~\ref{tab:Seg_table}).

\subsection{LAION Results}

Scaling is an important concept in computer vision, but we have not seen this concept applied to 3D primitive
generation. To that end, we collect approx. 1.8M natural images from LAION-Aesthetic. We use a recent SOTA depth
estimation network~\cite{depth_anything_v2} to obtain depth maps, and make reasonable camera calibration assumptions to
lift a 3D point cloud from the depth map. In particular, we use the
Hypersim~\cite{roberts2021hypersimphotorealisticsyntheticdataset} module that predicts metric depth and use its camera
parameters to get the point cloud for each image, which is required for training our convex decomposition model. GT normals can be obtained using the image gradient method described in~\cite{Vavilala_2023_ICCV}, which requires point cloud input.
{\bf LAION is easier than NYUv2} as Table~\ref{tab:nyuv2vsLAION} shows (note the improved AbsRel), likely due to the much larger dataset and availability of easier scenes. NYUv2 scenes are complex cluttered indoor rooms.

Our depth AbsRel numbers on LAION (comparing primitive depth against supplied depth map) lie between .019 to .029 (see Table~\ref{tab:laion_6_full}). For reference, the state-of-the-art depth estimator DepthAnythingV2~\cite{depth_anything_v2} reports AbsRel values (predicted vs. ground truth depth) of approximately 0.044 to 0.075 on NYUv2 and KITTI, which helps contextualize that our depth errors are comparatively small.

{\bf Our network start is much better than pure descent}, as Fig.~\ref{fig:FineTune12} shows.  The randomly started pure
descent procedure of Section~\ref{polishing} produces surprisingly strong fits, but requires a large number of
iterations to do so.  Typically, 100 iterations of polishing a network start point is much better than 1000 iterations
of pure descent.  The descent procedure is a first order method, so we expect AbsRel to improve no faster than
1/iterations, suggesting that this figure understates the advantage of the network start point.

\section{Discussion}
\label{sec:disc}
\begin{figure}
  \centering
  \includegraphics[width=1.0\linewidth]{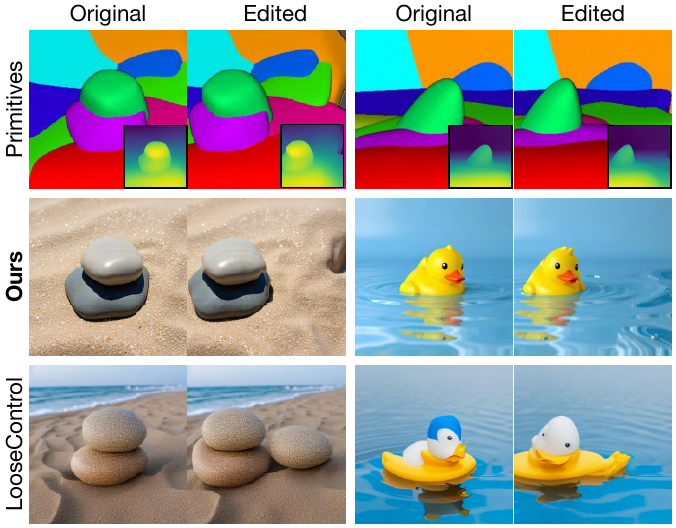}
  \caption{
Primitives support image editing. Figure reproduced from concurrent submission~\cite{vavilala2025generativeblocksworldmoving}. 
Primitives yield camera moves by a pipeline that: fits primitives to a single RGBD image from a scene;
uses those primitives to transfer image texture to an image from the new viewpoint, recording where textures are uncertain or unknown;
then synthesizes an image using a depth conditioned diffusion model with the transferred image as a hint. Examples
where objects are moved appear in supplementary~\cite{vavilala2025generativeblocksworldmoving}. This figure 
compares our primitives with results from LooseControl \protect \cite{bhat2023loosecontrol}.
Figures show scenes, primitives and depth maps, where "Edited" is post camera move. In each case, the camera is moved to the right.
LooseControl is at a disadvantage in this task, because its primitives, while compact, are intentionally not
accurate, so the duck changes and an extra rock appears. In each case, our primitive
representation  is accurate enough to show the same scene from a different view.
}
  \label{fig:b2w_2}
\end{figure}

Primitives are an old obsession in computer vision.  Their original purpose (object recognition) now
appears to be much better handled in other ways.  Mostly, using primitives was never really an issue, because
there weren't viable fitting procedures.  But what are primitives for? Likely answers come from robotics -- where one
might benefit from simplified representations of geometry that are still accurate -- and image editing -- where a user
might edit a scene by moving primitives (e.g. Fig.~\ref{fig:b2w_2}). Future work may better handle variable primitive representations (we train a separate network for each primitive count $K^{total}/K^-$) and improve the network start, eliminating the need for post-training finetuning. While this paper has drastically advanced geometric accuracy of 3D primitive fitting to data, improving segmentation boundaries remains open.

\section{Acknowledgments}
This material is based upon work supported by the National Science Foundation under Grant No. 2106825 and by a gift from the Boeing Corporation. This research used the Delta advanced computing and data resource which is supported by the National Science Foundation (award OAC 2005572) and the State of Illinois.

\clearpage
\newpage
{
    \small
    \bibliographystyle{ieeenat_fullname}
    \bibliography{main-daf}

@String(IJCV = {Int. J. Comput. Vis.})

@String(CVPR= {IEEE Conf. Comput. Vis. Pattern Recog.})

@String(ICCV= {Int. Conf. Comput. Vis.})

@String(ECCV= {Eur. Conf. Comput. Vis.})

@String(ICPR = {Int. Conf. Pattern Recog.})

@String(TOG= {ACM Trans. Graph.})

@String(AAAI = {AAAI})

@String(IJCV  = {IJCV})

@String(CVPR  = {CVPR})

@String(ICCV  = {ICCV})

@String(ECCV  = {ECCV})

@String(ICPR  = {ICPR})

@String(TOG   = {ACM TOG})

@article{monnier2023differentiable,
  title={Differentiable blocks world: Qualitative 3d decomposition by rendering primitives},
  author={Monnier, Tom and Austin, Jake and Kanazawa, Angjoo and Efros, Alexei and Aubry, Mathieu},
  journal={Advances in Neural Information Processing Systems},
  volume={36},
  pages={5791--5807},
  year={2023}
}

@inproceedings{alaniz2023iterative,
  title={Iterative superquadric recomposition of 3d objects from multiple views},
  author={Alaniz, Stephan and Mancini, Massimiliano and Akata, Zeynep},
  booktitle={Proceedings of the IEEE/CVF International Conference on Computer Vision},
  pages={18013--18023},
  year={2023}
}

@inproceedings{uy2022point2cyl,
  title={Point2Cyl: Reverse Engineering 3D Objects From Point Clouds to Extrusion Cylinders},
  author={Uy, Mikaela Angelina and Chang, Yen-Yu and Sung, Minhyuk and Goel, Purvi and Lambourne, Joseph G and Birdal, Tolga and Guibas, Leonidas J},
booktitle = {CVPR},
  year={2022}
}

@inproceedings{wu2021deepcad,
  title={Deepcad: A deep generative network for computer-aided design models},
  author={Wu, Rundi and Xiao, Chang and Zheng, Changxi},
  booktitle={ICCV},
  year={2021}
}

@inproceedings{paschalidou2021neural,
  title={Neural parts: Learning expressive 3d shape abstractions with invertible neural networks},
  author={Paschalidou, Despoina and Katharopoulos, Angelos and Geiger, Andreas and Fidler, Sanja},
  booktitle={CVPR},
  year={2021}
}

@inproceedings{yu2022capri,
  title={Capri-net: Learning compact cad shapes with adaptive primitive assembly},
  author={Yu, Fenggen and Chen, Zhiqin and Li, Manyi and Sanghi, Aditya and Shayani, Hooman and Mahdavi-Amiri, Ali and Zhang, Hao},
  booktitle={Proceedings of the IEEE/CVF Conference on Computer Vision and Pattern Recognition},
  pages={11768--11778},
  year={2022}
}

@article{kluger2024robust,
  title={Robust Shape Fitting for 3D Scene Abstraction},
  author={Kluger, Florian and Brachmann, Eric and Yang, Michael Ying and Rosenhahn, Bodo},
  journal={IEEE Transactions on Pattern Analysis and Machine Intelligence},
  year={2024},
  publisher={IEEE}
}

@inproceedings{deng2019cerberus,
  title={Cerberus: A multi-headed derenderer},
  author={Deng, Boyang and Kornblith, Simon and Hinton, Geoffrey},
  maintitle = {The IEEE/CVF Conference on Computer Vision and Pattern Recognition (CVPR)},
  booktitle = {Workshop on 3D Scene Understanding},
  year={2019},
}

@article{fureview,
author={K. Fu and J. Peng and Q. He  et al},
title={Single image 3D object reconstruction based on deep learning: A review},
journal={Multimed Tools Appl},
volume=80,
pages={463–498},
year=2021,
}

@InProceedings{Vavilala_2023_ICCV,
    author    = {Vavilala, Vaibhav and Forsyth, David},
    title     = {Convex Decomposition of Indoor Scenes},
    booktitle = {Proceedings of the IEEE/CVF International Conference on Computer Vision (ICCV)},
    month     = {October},
    year      = {2023},
    pages     = {9176-9186}
}

@article{spaghetti,
title={SPAGHETTI: editing implicit shapes through part aware generation},
author={A. Hertz and O. Perel and O. Sorkine-Hornung and D. Cohen-Or},
journal={ACM Transactions on Graphics},
volume=41,
number=4,
pages={1-20},
year={2022},
}

@article{weiacd,
title={Approximate convex decomposition for 3D meshes with collision-aware concavity and tree search},
author={X. Wei and M. Liu and Z. Ling and H. Su},
journal={ACM Transactions on Graphics},
volume={41},
number=4,
year={2022},
}

@article{Calderon2017BoundingPF,
  title={Bounding proxies for shape approximation},
  author={St{\'e}phane Calderon and Tamy Boubekeur},
  journal={ACM Transactions on Graphics (TOG)},
  year={2017},
  volume={36},
  pages={1 - 13}
}

@article{Mo2019StructureNetHG,
  title={StructureNet: Hierarchical Graph Networks for 3D Shape Generation},
  author={Kaichun Mo and Paul Guerrero and L. Yi and Hao Su and Peter Wonka and Niloy Jyoti Mitra and Leonidas J. Guibas},
  journal={ACM Trans. Graph.},
  year={2019},
  volume={38},
  pages={242:1-242:19}
}

@article{Gadelha2020LearningGM,
  title={Learning Generative Models of Shape Handles},
  author={Matheus Gadelha and Giorgio Gori and Duygu Ceylan and Radom{\'i}r Mech and Nathan A. Carr and Tamy Boubekeur and Rui Wang and Subhransu Maji},
  journal={2020 IEEE/CVF Conference on Computer Vision and Pattern Recognition (CVPR)},
  year={2020},
  pages={399-408}
}

@misc{bhat2023loosecontrol,
      title={LooseControl: Lifting ControlNet for Generalized Depth Conditioning}, 
      author={Shariq Farooq Bhat and Niloy J. Mitra and Peter Wonka},
      year={2023},
      eprint={2312.03079},
      archivePrefix={arXiv},
      primaryClass={cs.CV}
}

@article{Roberts2021LSDStructureNetML,
  title={LSD-StructureNet: Modeling Levels of Structural Detail in 3D Part Hierarchies},
  author={Dominic Roberts and Aram Danielyan and Hang Chu and Mani Golparvar Fard and David A. Forsyth},
  journal={2021 IEEE/CVF International Conference on Computer Vision (ICCV)},
  year={2021},
  pages={5816-5825}
}

@article{Smirnov2019DeepPS,
  title={Deep Parametric Shape Predictions Using Distance Fields},
  author={Dmitriy Smirnov and Matthew Fisher and Vladimir G. Kim and Richard Zhang and Justin M. Solomon},
  journal={2020 IEEE/CVF Conference on Computer Vision and Pattern Recognition (CVPR)},
  year={2019},
  pages={558-567}
}

@article{Sun2019LearningAH,
  title={Learning adaptive hierarchical cuboid abstractions of 3D shape collections},
  author={Chun-Yu Sun and Qian-Fang Zou},
  journal={ACM Transactions on Graphics (TOG)},
  year={2019},
  volume={38},
  pages={1 - 13}
}

@article{Barr1981SuperquadricsAA,
  title={Superquadrics and Angle-Preserving Transformations},
  author={Barr},
  journal={IEEE Computer Graphics and Applications},
  year={1981},
  volume={1},
  pages={11-23}
}

@inproceedings{Jakli2000SegmentationAR,
  title={Segmentation and Recovery of Superquadrics},
  author={Ale{\c{s}} Jakli{\v{c}} and Ale{\v{s}} Leonardis and Franc Solina},
  booktitle={Computational Imaging and Vision},
  year={2000}
}

@article{Paschalidou2019SuperquadricsRL,
  title={Superquadrics Revisited: Learning 3D Shape Parsing Beyond Cuboids},
  author={Despoina Paschalidou and Ali O. Ulusoy and Andreas Geiger},
  journal={2019 IEEE/CVF Conference on Computer Vision and Pattern Recognition (CVPR)},
  year={2019},
  pages={10336-10345}
}

@article{Zou20173DPRNNGS,
  title={3D-PRNN: Generating Shape Primitives with Recurrent Neural Networks},
  author={Chuhang Zou and Ersin Yumer and Jimei Yang and Duygu Ceylan and Derek Hoiem},
  journal={2017 IEEE International Conference on Computer Vision (ICCV)},
  year={2017},
  pages={900-909}
}

@article{MarrNishihara1978,
  author = {Marr, D. and Nishihara, H. K.},
  title = {Representation and recognition of the spatial organization of three-dimensional shapes},
  journal = {Proceedings of the Royal Society of London. Series B. Biological Sciences},
  volume = {200},
  number = {1140},
  pages = {269--294},
  year = {1978},
  doi = {10.1098/rspb.1978.0020},
  publisher = {Royal Society},
  issn = {0080-4649},
  url = {https://doi.org/10.1098/rspb.1978.0020}
}

@article{Li2018SupervisedFO,
  title={Supervised Fitting of Geometric Primitives to 3D Point Clouds},
  author={Lingxiao Li and Minhyuk Sung and Anastasia Dubrovina and L. Yi and Leonidas J. Guibas},
  journal={2019 IEEE/CVF Conference on Computer Vision and Pattern Recognition (CVPR)},
  year={2018},
  pages={2647-2655}
}

@misc{vavilala2025generativeblocksworldmoving,
      title={Generative Blocks World: Moving Things Around in Pictures}, 
      author={Vaibhav Vavilala and Seemandhar Jain and Rahul Vasanth and D. A. Forsyth and Anand Bhattad},
      year={2025},
      eprint={2506.20703},
      archivePrefix={arXiv},
      primaryClass={cs.GR},
      url={https://arxiv.org/abs/2506.20703}, 
}

@inproceedings{Ramamonjisoa2022MonteBoxFinderDA,
  title={MonteBoxFinder: Detecting and Filtering Primitives to Fit a Noisy Point Cloud},
  author={Ramamonjisoa, Micha{\"e}l and Stekovic, Sinisa and Lepetit, Vincent},
  booktitle={Computer Vision -- ECCV 2022},
  pages={160--176},
  year={2022},
  publisher={Springer},
  series={Lecture Notes in Computer Science},
  volume={13688}
}

@article{Hampali2021MonteCS,
  title={Monte Carlo Scene Search for 3D Scene Understanding},
  author={Shreyas Hampali and Sinisa Stekovic and Sayan Deb Sarkar and Chetan Srinivasa Kumar and Friedrich Fraundorfer and Vincent Lepetit},
  journal={2021 IEEE/CVF Conference on Computer Vision and Pattern Recognition (CVPR)},
  year={2021},
  pages={13799-13808}
}

@inproceedings{kluger2020consac,
	title        = {{CONSAC}: {R}obust {M}ulti-{M}odel {F}itting by {C}onditional {S}ample {C}onsensus},
	author       = {Florian Kluger and Eric Brachmann and Hanno Ackermann and Carsten Rother and Michael Ying Yang and Bodo Rosenhahn},
	year         = 2020,
	booktitle    = {CVPR}
}

@inproceedings{kluger2024parsac,
  title={{PARSAC}: {A}ccelerating {R}obust {M}ulti-{M}odel {F}itting with {P}arallel {S}ample {C}onsensus},
  author={Kluger, Florian and Rosenhahn, Bodo},
  booktitle={AAAI},
  year={2024}
}

@article{Chen2019BSPNetGC,
  title={BSP-Net: Generating Compact Meshes via Binary Space Partitioning},
  author={Zhiqin Chen and Andrea Tagliasacchi and Hao Zhang},
  journal={2020 IEEE/CVF Conference on Computer Vision and Pattern Recognition (CVPR)},
  year={2019},
  pages={42-51}
}

@article{Kang2020ARO,
  title={A Review of Techniques for 3D Reconstruction of Indoor Environments},
  author={Zhizhong Kang and Juntao Yang and Zhou Yang and Sai Cheng},
  journal={ISPRS Int. J. Geo Inf.},
  year={2020},
  volume={9},
  pages={330}
}

@article{Liu2018PlaneRCNN3P,
  title={PlaneRCNN: 3D Plane Detection and Reconstruction From a Single Image},
  author={Chen Liu and Kihwan Kim and Jinwei Gu and Yasutaka Furukawa and Jan Kautz},
  journal={2019 IEEE/CVF Conference on Computer Vision and Pattern Recognition (CVPR)},
  year={2018},
  pages={4445-4454}
}

@inproceedings{kluger2021cuboids,
  title={Cuboids Revisited: Learning Robust 3D Shape Fitting to Single RGB Images},
  author={Kluger, Florian and Ackermann, Hanno and Brachmann, Eric and Yang, Michael Ying and Rosenhahn, Bodo},
  booktitle={Proceedings of the IEEE Conference on Computer Vision and Pattern Recognition (CVPR)},
  year={2021}
}

@article{deng2020cvxnet,
  title = {CvxNet: Learnable Convex Decomposition},
  author = {Deng, Boyang and Genova, Kyle and Yazdani, Soroosh and Bouaziz, Sofien and Hinton, Geoffrey and Tagliasacchi, Andrea},
  booktitle = {The IEEE/CVF Conference on Computer Vision and Pattern Recognition (CVPR)},
  month = {June},
  year = {2020},
}

@article{Hoiem:2005:PopUp,
author = "Derek Hoiem and Alexei A. Efros and Martial Hebert",
title = "Automatic Photo Pop-up",
year = "2005",
month = aug,
journal = "ACM Transactions on Graphics / SIGGRAPH",
volume = "24",
number = "3",
}

@inproceedings{Silberman:ECCV12,
  author    = {Nathan Silberman, Derek Hoiem, Pushmeet Kohli and Rob Fergus},
  title     = {Indoor Segmentation and Support Inference from RGBD Images},
  booktitle = {ECCV},
  year      = {2012}
}

@inproceedings{liu2022towards,
  title={Towards High-Fidelity Single-View Holistic Reconstruction of Indoor Scenes},
  author={Liu, Haolin and Zheng, Yujian and Chen, Guanying and Cui, Shuguang and Han, Xiaoguang},
  booktitle={European Conference on Computer Vision},
  pages={429--446},
  year={2022},
  organization={Springer}
}

@inproceedings{liu2018planenet,
  title={Planenet: Piece-wise planar reconstruction from a single rgb image},
  author={Liu, Chen and Yang, Jimei and Ceylan, Duygu and Yumer, Ersin and Furukawa, Yasutaka},
  booktitle={Proceedings of the IEEE Conference on Computer Vision and Pattern Recognition},
  pages={2579--2588},
  year={2018}
}

@inproceedings{stekovic2020general,
  title={General 3d room layout from a single view by render-and-compare},
  author={Stekovic, Sinisa and Hampali, Shreyas and Rad, Mahdi and Sarkar, Sayan Deb and Fraundorfer, Friedrich and Lepetit, Vincent},
  booktitle={European Conference on Computer Vision},
  pages={187--203},
  year={2020},
  organization={Springer}
}

@inproceedings{jiang2014finding,
  title={Finding approximate convex shapes in rgbd images},
  author={Jiang, Hao},
  booktitle={European Conference on Computer Vision},
  pages={582--596},
  year={2014},
  organization={Springer}
}

@inproceedings{wang2015designing,
  title={Designing deep networks for surface normal estimation},
  author={Wang, Xiaolong and Fouhey, David and Gupta, Abhinav},
  booktitle={Proceedings of the IEEE conference on computer vision and pattern recognition},
  pages={539--547},
  year={2015}
}

@inProceedings{abstractionTulsiani17,
  title={Learning Shape Abstractions by Assembling Volumetric Primitives},
  author = {Shubham Tulsiani and Hao Su and Leonidas J. Guibas and Alexei A. Efros and Jitendra Malik},
  booktitle={Computer Vision and Pattern Regognition (CVPR)},
  year={2017}
}

@article{Tatarchenko2019WhatDS,
  title={What Do Single-View 3D Reconstruction Networks Learn?},
  author={Maxim Tatarchenko and Stephan R. Richter and Ren{\'e} Ranftl and Zhuwen Li and Vladlen Koltun and Thomas Brox},
  journal={2019 IEEE/CVF Conference on Computer Vision and Pattern Recognition (CVPR)},
  year={2019},
  pages={3400-3409}
}

@inproceedings{depthanything,
  title={Depth Anything: Unleashing the Power of Large-Scale Unlabeled Data},
  author={Yang, Lihe and Kang, Bingyi and Huang, Zilong and Xu, Xiaogang and Feng, Jiashi and Zhao, Hengshuang},
  booktitle={CVPR},
  year={2024}
}

@article{Ranftl2020,
	author    = {Ren\'{e} Ranftl and Katrin Lasinger and David Hafner and Konrad Schindler and Vladlen Koltun},
	title     = {Towards Robust Monocular Depth Estimation: Mixing Datasets for Zero-shot Cross-dataset Transfer},
	journal   = {IEEE Transactions on Pattern Analysis and Machine Intelligence (TPAMI)},
	year      = {2020},
}

@inproceedings{binford71,
author={TO Binford},
title={Visual perception by computer},
booktitle={IEEE Conf. on Systems and Controls},
year=1971,
}

@article{biederman,
author={I Biederman},
title={Recognition by components : A theory of human image understanding},
journal={Psychological Review},
number=94,
pages={115-147},
year=1987,
}

@inproceedings{shgc,
author={S. Shafer and T. Kanade},
title={The theory of straight homogeneous generalized cylinders},
booktitle={Technical Report CS-083-105, Carnegie Mellon University},
year=1983,
}

@Article{	  nevatia77,
  author	= {R. Nevatia and T.O. Binford},
  journal	= {Artificial Intelligence},
  title		= {Description and recognition of complex curved objects},
  year		= {1977}
}

@inproceedings{hebertponce,
author={J. Ponce and M. Hebert},
title={A new method for segmenting 3-d scenes into primitives},
booktitle={Proc. 6 ICPR},
year=1982,
}

@inproceedings{Fouhey13,
  title = {Data-Driven {3D} Primitives for Single Image Understanding},
  author = {Fouhey, David F. and Gupta, Abhinav and Hebert, Martial},
  booktitle = {ICCV},
  year = {2013},
}

@InProceedings{	  hedauiccv2009,
  author	= {V. Hedau and D. Hoiem and D. Forsyth},
  booktitle	= {Proc. ICCV},
  title		= {{Recovering the Spatial Layout of Cluttered Rooms}},
  year		= 2009
}

@InProceedings{	  hedaucvpr2012,
  author	= {V. Hedau and D. Hoiem and D. Forsyth},
  booktitle	= {Proc. CVPR},
  title		= {{Recovering Free Space of Indoor Scenes from a Single
		  Image}},
  year		= 2012
}

@InProceedings{	  hedaueccv2010,
  author	= {Varsha Hedau and Derek Hoiem and David Forsyth},
  booktitle	= {Proc. ECCV},
  title		= {{Thinking Inside the Box: Using Appearance Models and
		  Context Based on Room Geometry}},
  year		= 2010
}

@InProceedings{Zou_2018_CVPR,
author = {Zou, Chuhang and Colburn, Alex and Shan, Qi and Hoiem, Derek},
title = {LayoutNet: Reconstructing the 3D Room Layout From a Single RGB Image},
booktitle = {Proceedings of the IEEE Conference on Computer Vision and Pattern Recognition (CVPR)},
month = {June},
year = {2018}
}

@InProceedings{Zou_2017_ICCV,
author = {Zou, Chuhang and Yumer, Ersin and Yang, Jimei and Ceylan, Duygu and Hoiem, Derek},
title = {3D-PRNN: Generating Shape Primitives With Recurrent Neural Networks},
booktitle = {Proceedings of the IEEE International Conference on Computer Vision (ICCV)},
month = {Oct},
year = {2017}
}

@Article{	  hoiemijcv2007,
  author	= {D. Hoiem and A.~A. Efros and M. Hebert},
  journal	= {IJCV},
  title		= {{Recovering surface layout from an image}},
  year		= {2007}
}

@InProceedings{	  s:gupta10,
  author	= {Abhinav Gupta and Alexei A. Efros and Martial Hebert},
  title		= {Blocks World Revisited: Image Understanding using
		  Qualitative Geometry and Mechanics},
  booktitle	= eccv,
  year		= {2010}
}

@inproceedings{lego,
author={A. van den Hengel and C. Russell and A. Dick and J. Bastian and D. Poo-
ley, L. Fleming and L. Agapito},
title={Part-based modelling of
compound scenes from images},
booktitle={CVPR},
year={2015},
}

@book{roberts,
author={L. G. Roberts},
title={Machine Perception of Three-Dimensional
Solids},
publisher={PhD thesis, MIT},
year=1963,
}

@inproceedings{
jang2017categorical,
title={Categorical Reparameterization with Gumbel-Softmax},
author={Eric Jang and Shixiang Gu and Ben Poole},
booktitle={International Conference on Learning Representations},
year={2017},
url={https://openreview.net/forum?id=rkE3y85ee}
}

@article{ransac,
author={M. A. Fischler and R. C. Bolles},
year=1981,
title={Random Sample Consensus: A Paradigm for Model Fitting with Applications to Image Analysis and Automated Cartography},
journal={Comm. ACM.},
volume=24,
number=6,
pages={381–395},
}

@ARTICLE {Ranftl2022,
    author  = "Ren\'{e} Ranftl and Katrin Lasinger and David Hafner and Konrad Schindler and Vladlen Koltun",
    title   = "Towards Robust Monocular Depth Estimation: Mixing Datasets for Zero-Shot Cross-Dataset Transfer",
    journal = "IEEE Transactions on Pattern Analysis and Machine Intelligence",
    year    = "2022",
    volume  = "44",
    number  = "3"
}

@misc{loshchilov2019decoupledweightdecayregularization,
      title={Decoupled Weight Decay Regularization}, 
      author={Ilya Loshchilov and Frank Hutter},
      year={2019},
      eprint={1711.05101},
      archivePrefix={arXiv},
      primaryClass={cs.LG},
      url={https://arxiv.org/abs/1711.05101}, 
}

@misc{roberts2021hypersimphotorealisticsyntheticdataset,
      title={Hypersim: A Photorealistic Synthetic Dataset for Holistic Indoor Scene Understanding}, 
      author={Mike Roberts and Jason Ramapuram and Anurag Ranjan and Atulit Kumar and Miguel Angel Bautista and Nathan Paczan and Russ Webb and Joshua M. Susskind},
      year={2021},
      eprint={2011.02523},
      archivePrefix={arXiv},
      primaryClass={cs.CV},
      url={https://arxiv.org/abs/2011.02523}, 
}

@inproceedings{depth_anything_v2,
  title={Depth Anything V2},
  author={Yang, Lihe and Kang, Bingyi and Huang, Zilong and Zhao, Zhen and Xu, Xiaogang and Feng, Jiashi and Zhao, Hengshuang},
  booktitle={Advances in Neural Information Processing Systems},
  volume={37},
  year={2024}
}

@inproceedings{Tawfik, author = {Tawfik, Maged S.}, title = {An efficient algorithm for CSG to b-rep conversion}, year = {1991}, isbn = {0897914279}, publisher = {Association for Computing Machinery}, address = {New York, NY, USA}, url = {https://doi.org/10.1145/112515.112534}, doi = {10.1145/112515.112534}, booktitle = {Proceedings of the First ACM Symposium on Solid Modeling Foundations and CAD/CAM Applications}, pages = {99–108}, numpages = {10}, location = {Austin, Texas, USA}, series = {SMA '91} }

@inproceedings{amatoprim,
title={Fast collision detection for motion planning using shape primitive skeletons},
author={M. Ghosh and S. Thomas and N. M. Amato},
year=2018,
booktitle={International Workshop on the Algorithmic Foundations of Robotics},
pages={36-51},
publisher={Springer International Publishing},}

@INPROCEEDINGS{sqgrasp,
  author={Vezzani, Giulia and Pattacini, Ugo and Pasquale, Giulia and Natale, Lorenzo},
  booktitle={2018 IEEE International Conference on Robotics and Automation (ICRA)}, 
  title={Improving Superquadric Modeling and Grasping with Prior on Object Shapes}, 
  year={2018},
  volume={},
  number={},
  pages={6875-6882},
  doi={10.1109/ICRA.2018.8463161}}

@inproceedings{primgrasp,
  title     = {Shape-Primitive Based Object Recognition and Grasping},
  author    = {Nieuwenhuisen, Matthias and Stückler, Jörg and Berner, Alexander and Klein, Reinhard and Behnke, Sven},
  booktitle = {Proceedings of the 7th German Conference on Robotics (ROBOTIK)},
  address   = {Munich, Germany},
  publisher = {VDE Verlag},
  year      = {2012},
  pages     = {1--5}
}

@inproceedings{primgrasp2,
author={A. T. Miller and S. Knoop and H. I. Christensen and P. K. Allen},
title={Automatic grasp planning using shape primitives},
year=2003,
booktitle={IEEE International Conference on Robotics and Automation},
}

@misc{tracy2023differentiablecollisiondetectionset,
      title={Differentiable Collision Detection for a Set of Convex Primitives}, 
      author={Kevin Tracy and Taylor A. Howell and Zachary Manchester},
      year={2023},
      eprint={2207.00669},
      archivePrefix={arXiv},
      primaryClass={cs.RO},
      url={https://arxiv.org/abs/2207.00669}, 
}

@misc{xu2025rgbsqgraspinferringlocalsuperquadric,
      title={RGBSQGrasp: Inferring Local Superquadric Primitives from Single RGB Image for Graspability-Aware Bin Picking}, 
      author={Yifeng Xu and Fan Zhu and Ye Li and Sebastian Ren and Xiaonan Huang and Yuhao Chen},
      year={2025},
      eprint={2503.02387},
      archivePrefix={arXiv},
      primaryClass={cs.RO},
      url={https://arxiv.org/abs/2503.02387}, 
}
}
\clearpage
\newpage

\setcounter{page}{1}
\maketitlesupplementary

\section{Optimizing the Inference Pipeline}
\label{sec:opt}

Given the computational cost of ensembling, we seek to maximize throughput of our inference pipeline. We use $\texttt{torch.jit}$ and pure
BFloat16 for encoding the RGBD image and finetuning. We also get speedups from batching the test images instead of one
at a time. Combined with our subsampling strategy, these improvements yield over an order of magnitude faster inference than prior work, making ensembling more practical (see Table~\ref{tab:Seg_table}).  

Since our primitives are the blended union of half-spaces~\cite{deng2020cvxnet}, they cannot be rasterized easily and raymarching the SDF is required.
We note that rendering the primitives still requires FP32 precision to avoid unwanted artifacts. We accelerate our
raymarcher by advancing the step size by 0.8*SDF if it is greater than the minimum step size (we use $0.001$ for large-scale
metrics gathering, $0.0001$ for beauty renders). We cannot advance by the full SDF because it is an approximation of
how far the smoothed primitive boundary is. We apply interval halving at the intersection point to refine the estimate.

\section{Negative Primitives Theory}
\label{sec:theory}

Set differencing can result in very efficient representations.  Qualitative evidence was shown in Fig~\ref{fig:neg_intro}, in which we model a cube with a hole punched in it. Intuitively, one positive and one negative primitive are sufficient to model it perfectly (2 total primitives). Without CSG, approx. 5 primitives may be required, which is less parameter-efficient. Based on that, we can sketch a theoretical argument as to why having a vocabulary of mixed positive and negative primitives is expected to yield more accurate representations than the same number of positive-only primitives.

\subsection{CSG Representational Efficiency}

It is known that,  for a CSG model in 3D with $n$ distinct faces in the CSG tree, the resulting object model can have $O(n^3)$ faces~\cite{Tawfik}.
Relatively little appears to be known about the effect of the number of negative primitives on the complexity.
We show that, under the circumstances that apply here, there are geometries that admit  
short descriptions using negative primitives and have very much longer descriptions when only unions are allowed.
The bound is obtained by reasoning about what is required to encode the area of an object.

For a shape $S$, let $K_+(S)$ be the minimum description length using only positive primitives, and $K_{\pm}(S)$ be the minimum description length using mixed primitives. We claim that, under the circumstances that apply here, there are shapes such that $K_{\pm}(S) \ll K_+(S)$. The construction is straightforward.

{\bf Preamble:}  Primitives are smoothed polytopes as described in~\cite{deng2020cvxnet}.  There is some (very large) finite bound on
the number of faces.  These primitives are convex by construction (so gaussian curvature is non-negative), but are smooth.
The surface of these primitives consists of {\em flat} regions (where both normal and gaussian curvature are very close to zero),
{\em edge} regions, where gaussian curvature is close to zero but normal curvature may be large, and {\em vertex} regions, where there is considerable gaussian curvature.  These regions correspond to 2-faces, 1-faces and 0-faces in the underlying polytope.  The actual values of the curvature are not significant
for our purposes.

\begin{figure}
\centering
\includegraphics[width=\linewidth]{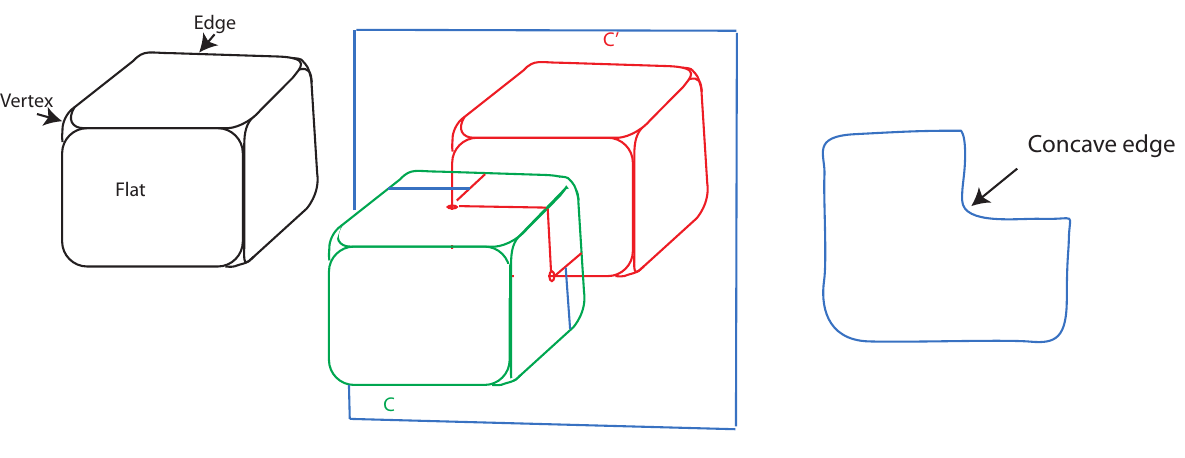}
\caption{A shape which cannot be efficiently encoded as a union of convexes can be built out of a smoothed unit cube (on the {\bf left},
  showing {\em flat}  regions, {\em edge} regions and {\em vertex} regions).  {\bf Center} sketches $C$ ({\bf green}) and $C'$ ({\bf red}),
  and a plane slicing the shape $C-C'$ ({\bf blue}).  {\bf Right} sketches the cross section cut by the plane: note the concave region, resulting from
  the smoothed convex edge region of $C'$.  This requires at least $O(1/\delta)$ convexes to approximate to precision $\delta$.}
\label{fig:cubeg}
\end{figure}

\newcommand{\Kpm}{K_{\pm}}
\newcommand{\Kp}{K_{+}}
\newcommand{\bigO}{O}
\newcommand{\bigOmega}{\Omega}
\newcommand{\calP}{\mathcal{P}}

\begin{theorem}[Restatement of Theorem~1 from main paper]
  For the vocabulary of primitives described,  there exist shapes $S$ such that $K_{\pm}(S) \ll K_+(S)$ when the representation of $S$ is
  required to achieve sufficiently high precision.
\end{theorem}

\subsection*{Sketch of Proof}

We construct a shape with this property.  Write $C$ for an approximation of a unit cube, represented as a smoothed polytope, and
$C'$ for that cube translated by $(0.5, 0.5, 0.5)$.  Now consider $S=C-C'$ (Figure~\ref{fig:cubeg}).  Clearly, $\Kpm(S)$ is $O(1)$.
Consider $\Kp(S)$ as the resolution $\delta$ of the representation increases.   Note from Figure~\ref{fig:cubeg} that representing
$S$ requires representing three concave edge-like regions, and one concave vertex-like region.  Since we can use only positive
primitives which are convex, we must use the flat faces.  It is straightforward that representing the concave edge-like regions to resolution
$\delta$ will require $O(1/\delta)$ flats, and representing the concave vertex-like region to resolution $\delta$ will require $O(1/\delta^2)$ flats.
It follows that
\[
\lim_{\delta\rightarrow 0} \frac{\Kpm}{\Kp}=0
\]
and so $\Kpm \ll \Kp$ in this case.  Notice the underlying geometry of the effect is common -- it takes a lot of convex shapes to represent
concave cutouts.  We expect that for ``most shapes'' the limit applies.

This theorem depends very delicately on the library of primitives and the transformations allowed to the primitives.
In some cases, it is hopelessly {\em optimistic}.  For example, if all primitives are cubes of two fixed sizes, where positives
are large and negatives small, and transformations are purely Euclidean, it may not be possible to encode a set difference
with positives only.  We are not aware of bounds that take these effects into account.

\subsection{Optimal Ratio of Negative Primitives in CSG Modeling}
A simple model leads to an intriguing result. For a fixed budget of primitives, we aim to determine the optimal ratio of negative primitives ($K^-$) to positive primitives ($K^+$) that maximizes representational efficiency. 

A CSG model can be described as: 
\begin{equation}
\text{Object} = (P_1 \cup P_2 \cup ...\cup P_{K^+}) - (N_1 \cup N_2 \cup ... \cup N_{K^-})
\end{equation}

Where $P_i$ are positive primitives and $N_j$ are negative primitives, with $K^+ + K^- = K^{total}$.

\textbf{Definitions}:
\begin{enumerate}
\item\textit{Primitive Interaction}: Overlapping volumes creating representational complexity
\item\textit{PP} Interaction: Between two positive primitives
\item\textit{PN} Interaction: Between a positive and negative primitive
\end{enumerate}

\textbf{Assumptions}:
\begin{enumerate}
    \item Only positive volumes and their modifications by negative primitives are visible in the final result
\item The representational power comes primarily from \textit{PP} and \textit{PN} interactions
\item Primitives are distributed to maximize meaningful interactions
\item Optimal representation maximizes visible features per primitive used
\item Assume connected geometry
\end{enumerate}

\textbf{Mathematical Model}:
\begin{enumerate}
\item Number of \textit{PP} Interactions: $\binom{K^+}{2} = \frac{K^+(K^+-1)}{2}$
\item Number of \textit{PN} Interactions: $K^+ \cdot K^-$
\end{enumerate}

\textbf{Balancing \textit{PP} and \textit{PN} Interactions:}

For optimal efficiency, \textit{PP} and \textit{PN} interactions should be balanced:
\begin{equation}
\frac{K^+(K^+-1)}{2} \approx K^+ \cdot K^-
\end{equation}

Substituting $K^+ = K^{total} - K^-$ and simplifying:
\begin{equation}
\frac{(K^{total} - K^-)(K^{total} - K^- - 1)}{2} \approx (K^{total} - K^-) \cdot K^- 
\end{equation}
\begin{equation}
\frac{K^{total} - K^- - 1}{2} \approx K^-
\end{equation}
For large $K^{total}$:
\begin{equation}
\frac{K^{total} - K^-}{2} \approx K^-
\end{equation}
Solving:
 \begin{equation}
     K^{total} \approx 3K^-
 \end{equation}
\begin{equation}
    \boxed{K^- \approx \frac{K^{total}}{3}}
\end{equation}

Thus, 
\begin{equation}
    \boxed{K^+ \approx \frac{2K^{total}}{3} }
\end{equation}

\textbf{Verification:} With this ratio, both interaction types equal approximately $\frac{2(K^{total})^2}{9}$, confirming our balance criterion.

\begin{figure*}[t!]
    \centering
    \pdfgrid{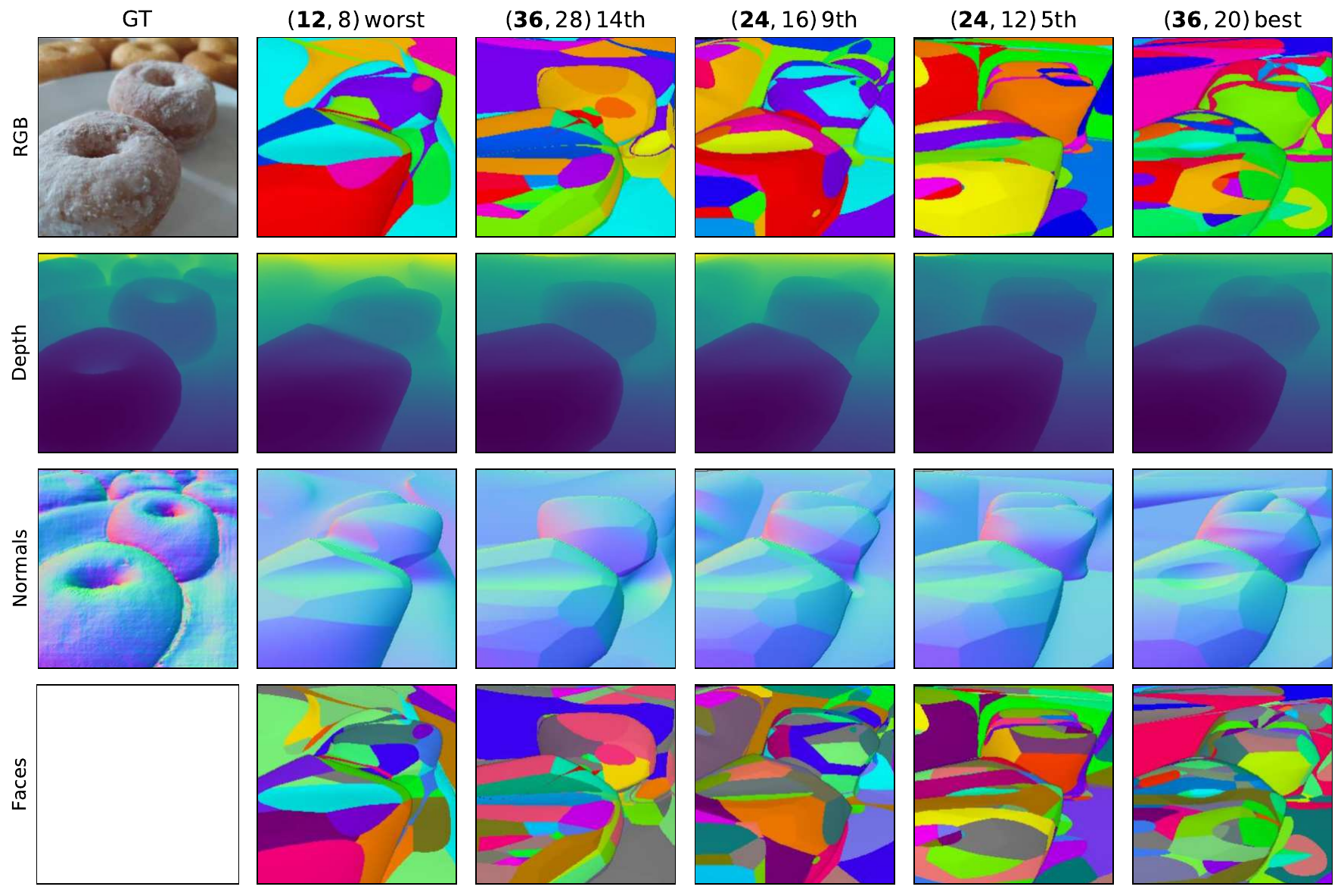}{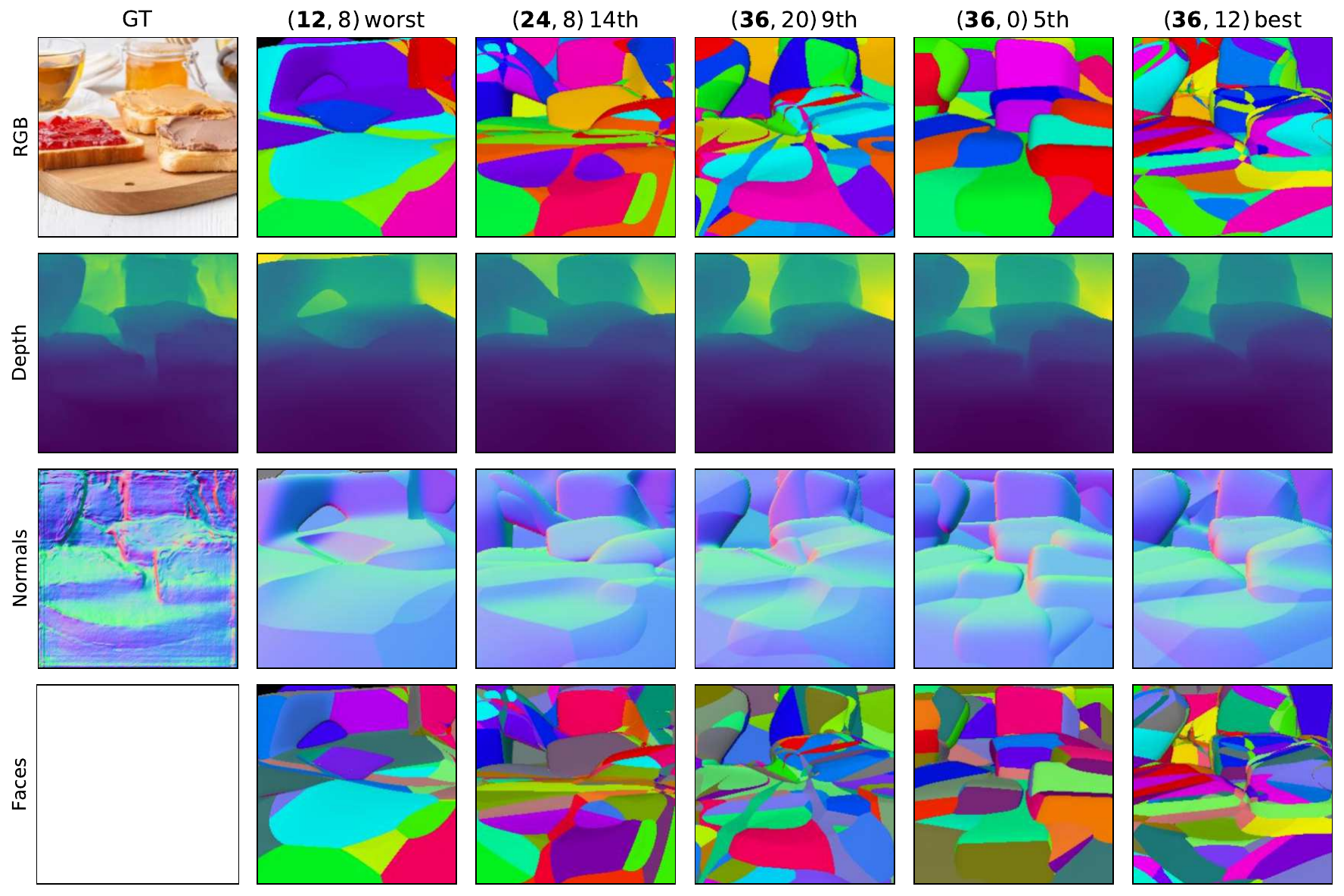}{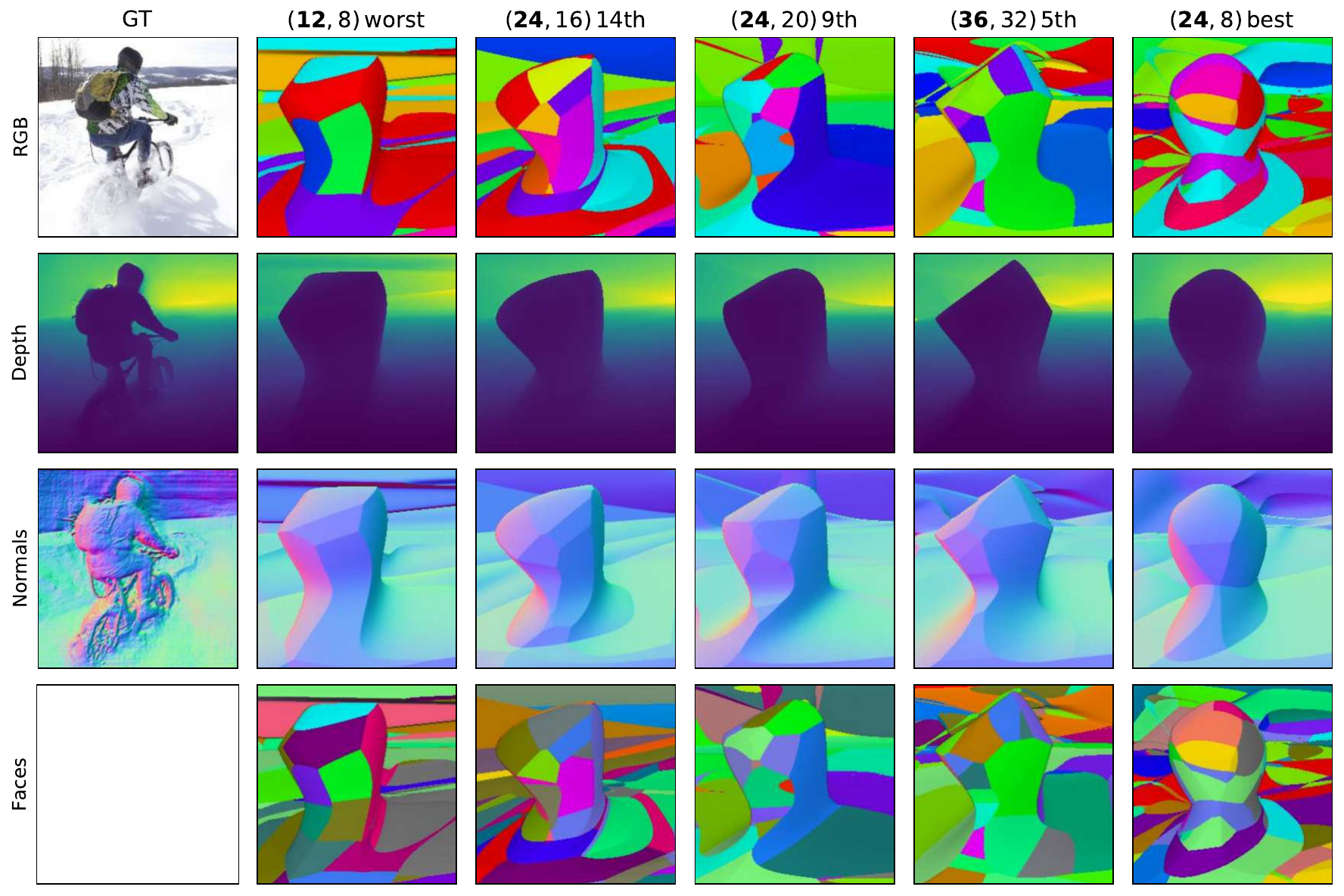}{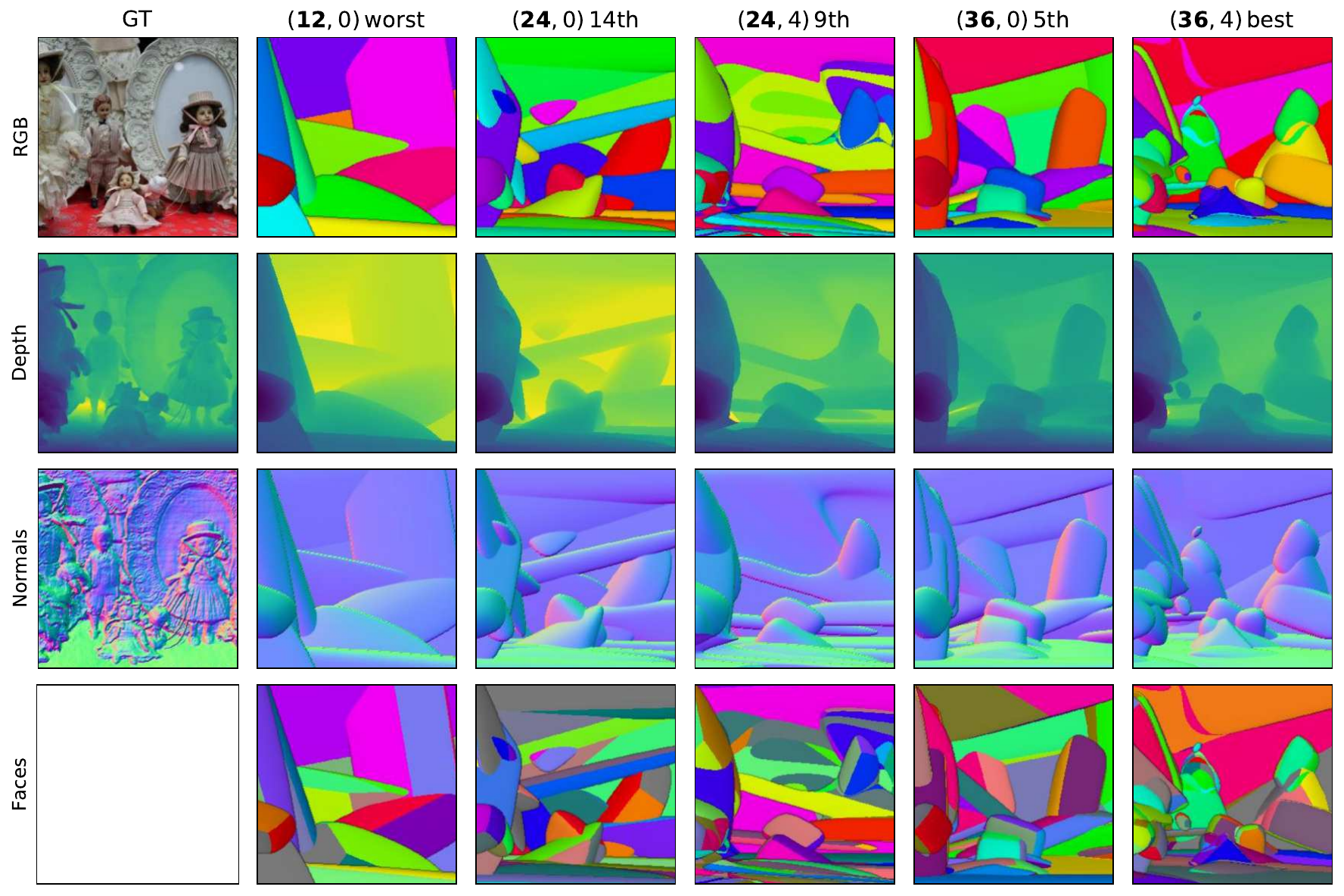}
    \caption{Additional qualitative examples shown.}
    \label{fig:quallaion2}
\end{figure*}

\textbf{Empirical Evidence and Practical Verification:}
This result is intriguing, because it is consistent with our experimental observations.   It should be noted that the assumptions may not be sound (the losses
on primitives should tend to lead to fewer interactions between positive primitives than our model requires),  but experimental results suggest quite strongly
that the best results are obtained when $K^{total}/K^-$ is about 3.   We believe this is likely some form of formal geometric property, rather than
a coincidence.  In our experimentation, all ratios of $K^{total}/K^-$ provide excellent primitive representations, showing that our underlying neural network, losses, and data pipeline are sound. But on average, the \textit{best} representations were near $K^{total}/K^- \approx 3:1$. Observe on both LAION and NYUv2 how depth and segmentation metrics tend to be best when $K^{total}/K^-$ are near $36/12$, $24/8$, and $12/4$ (Tables ~\ref{tab:nyu_full}, ~\ref{tab:laion_6_full}, ~\ref{tab:laion_12_full}).

\textbf{Limitations from Overlap Loss: } The theoretical model balances \textit{PP} interactions, representing the formation of the base positive volume, with \textit{PN} interactions, representing the carving of details. While the introduction of a loss encouraging positive primitives to spread out and avoid overlap alters the nature of \textit{PP} interactions---shifting them from forming dense, complex unions to defining a more distributed positive scaffold---the fundamental requirement for balancing these two generative forces remains. The term $\binom{K^+}{2}$ can thus be interpreted as the capacity of positive primitives to establish this initial, potentially dispersed, positive volume. Our quantitative evaluations, performed under these conditions, confirm that the optimal ratio of approximately $K^- \approx K^{total}/3$ persists. This empirical result suggests that the derived balance point robustly maximizes representational efficiency by ensuring sufficient primitives for both establishing the foundational positive elements of the scene and for subsequently sculpting them with negative primitives, aligning with principles of diminishing returns and complementary information.

\subsection{Face counts}
In Sec.~\ref{sec:boolean_primitives}, we described how the number of faces scales with the number of negative primitives. When computing segmentation accuracy with negative primitives, we compute the triple $(f_i,K^+_j,K^-_k)$ at each ray intersection point, where $i$ is the face index, $j$ is the index of the positive primitive we hit, and $k$ is the index of the (potentially) negative primitive we hit. Each unique triple can get its own face label. Thus, given a fixed primitive budget $K^{total}$, replacing a pure positive primitive representation with a mixture of positives and negatives can yield more unique faces. For example, $K^+/K^- = 12/0$ maxes out at $12f$ unique faces; $K^+/K^- = 6/6$ maxes out at $42f$ faces. Note that $f\times K^+ \times (1+K^-)$ is the theoretical maximum of unique labels, as practical scenes do not involve every primitive touching every other primitive.

\section{Primitives by Descent Alone}
    We generate a large
reservoir of 1M free-space (a.k.a. bbx samples) for each test image. We still generate $H\times W$ ``inside" surface
samples and ``outside" surface samples near the depth boundary respectively, with $\epsilon=0.02$ units separating these
surface samples. We remind the reader that our point clouds are renormalized to approx. the unit cube during training to
avoid scale issues. Then during finetuning, we subsample from all available samples at each step, providing a rich
gradient analogous to the network training process (though here, we're optimizing the parameters of primitives). We
found subsampling $10\%$ of available samples sufficient at each step.   

Second, we find that vanilla SGD does not produce usable results; instead
AdamW~\cite{loshchilov2019decoupledweightdecayregularization} was required. We set the initial LR to 0.01, and linearly
warm up to it over the first $25\%$ of iterations. We then halve the learning rate once at $50\%$ of the steps and again
at $75\%$.

\begin{figure*}
    \centering
    \pdfgrid{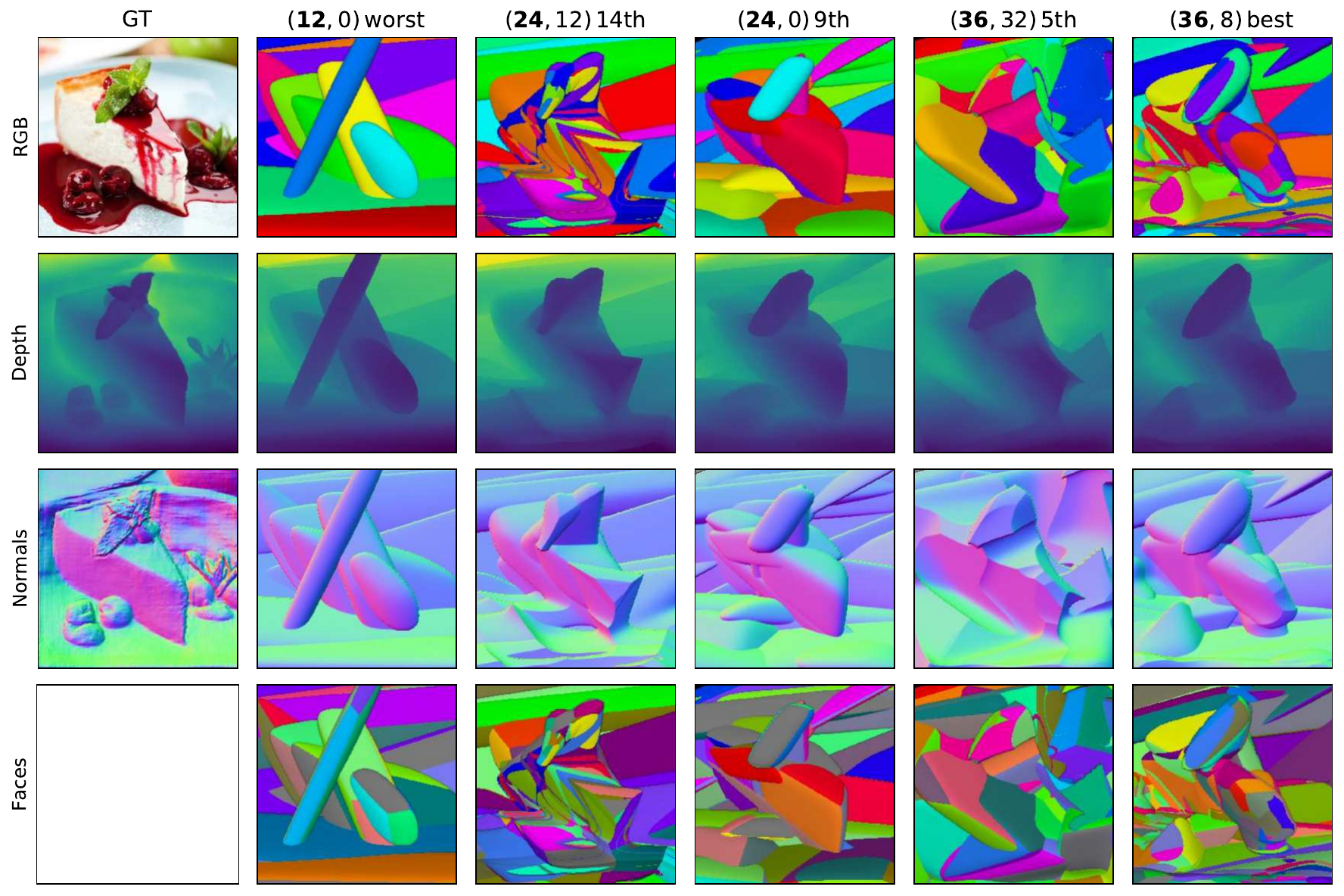}{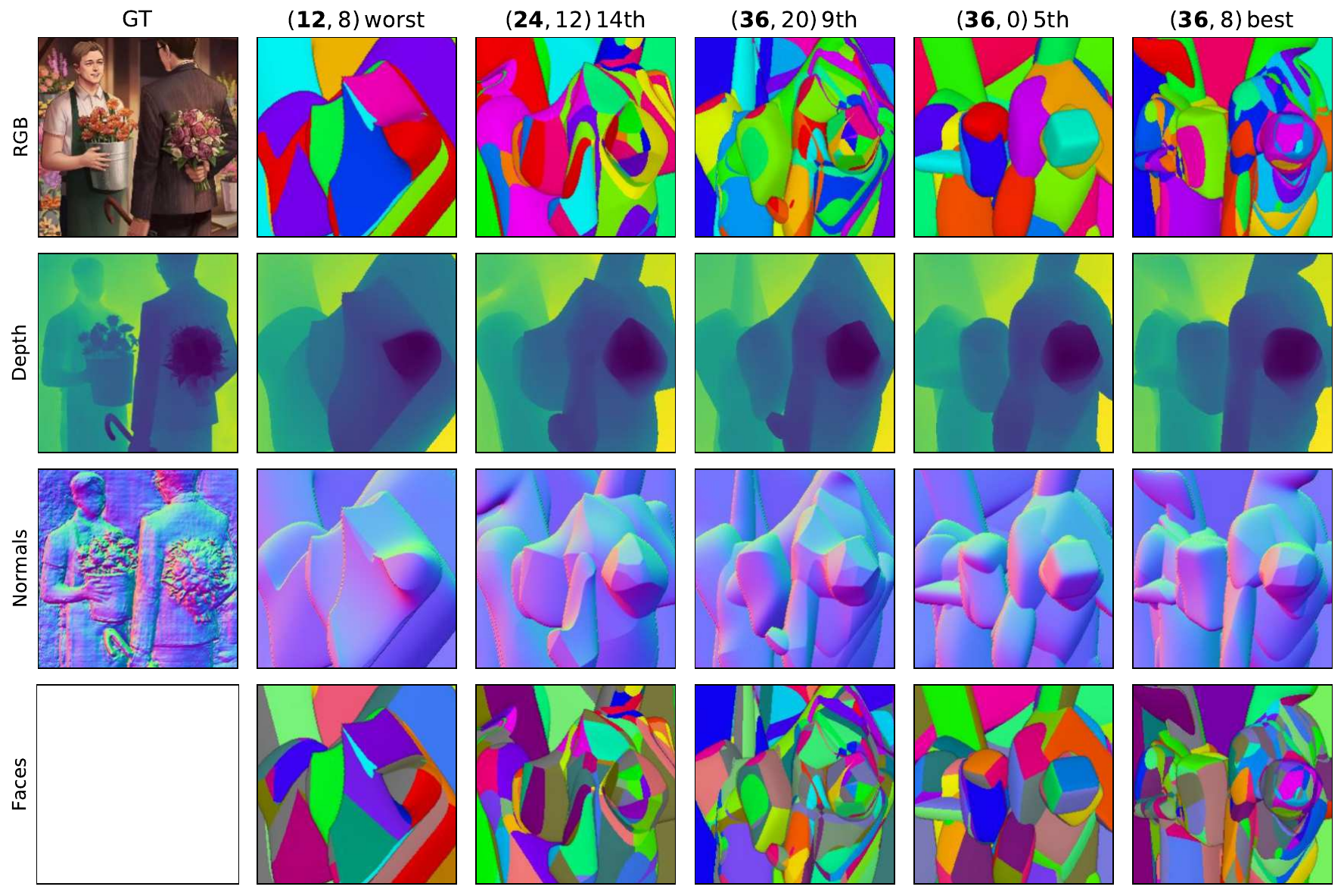}{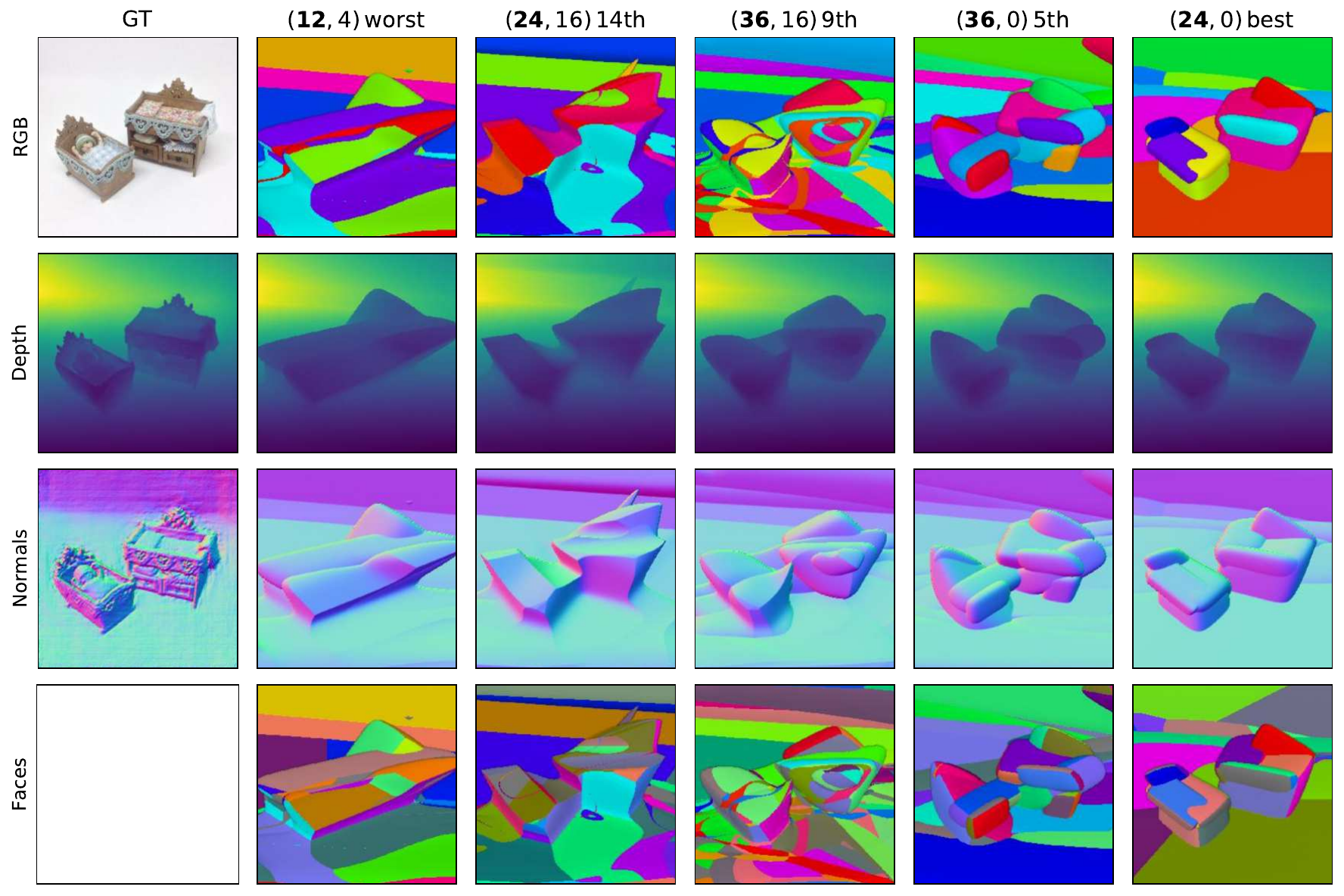}{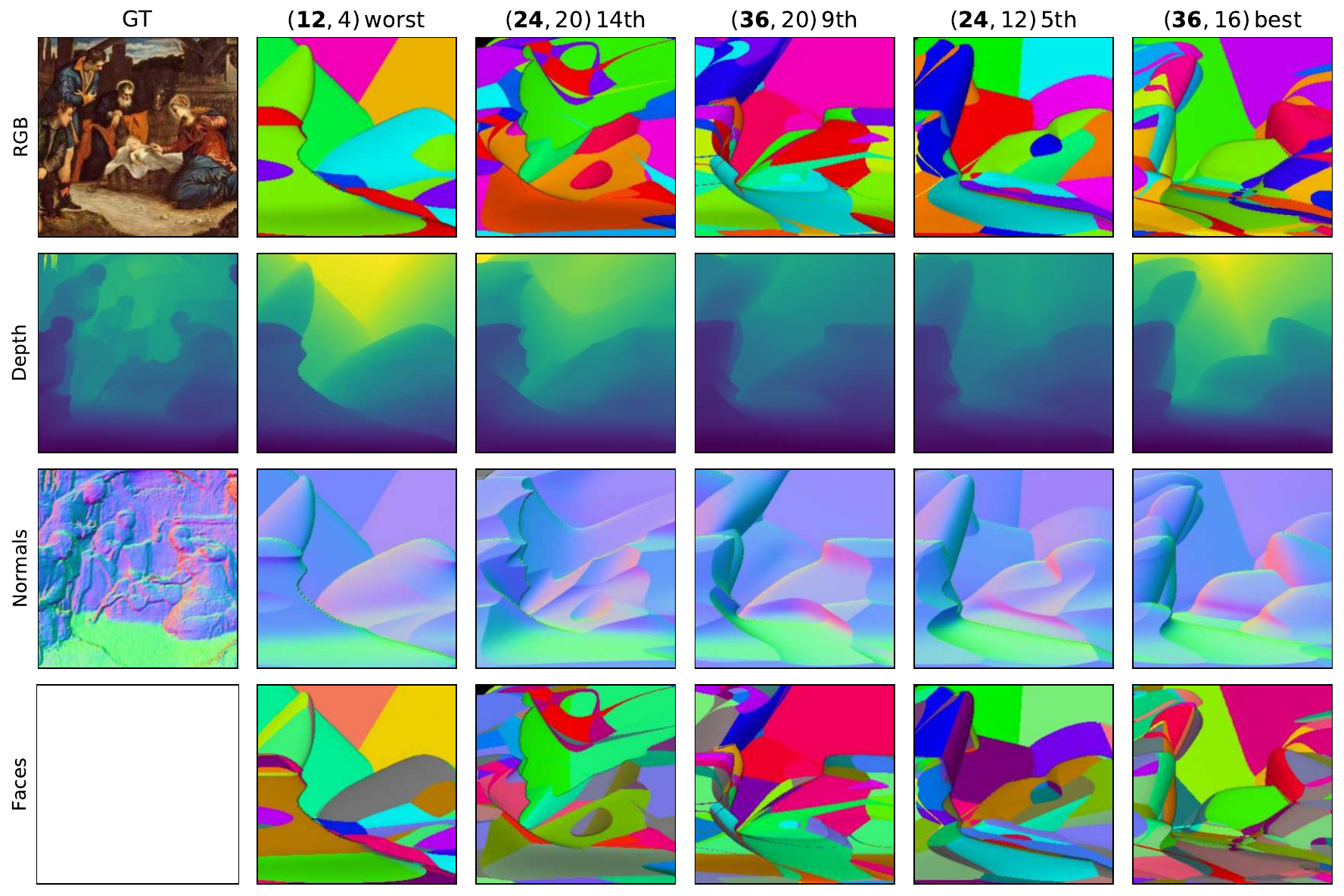}
    \caption{Additional qualitative examples shown.}
    \label{fig:quallaion3}
\end{figure*}

\begin{table*}[htbp]

\centering

\resizebox{\textwidth}{!}{%

\begin{tabular}{ccrrcccccc}
\toprule
Ensemble & Refine & $K^{total}$ & $K^-$ & $\text{AUC}_{@50}\uparrow$ & $\text{AUC}_{@20}\uparrow$ & $\text{AUC}_{@10}\uparrow$ & $\text{AUC}_{@5}\uparrow$ & $\text{mean}_{cm}\downarrow$ & $\text{median}_{cm}\downarrow$\\
\midrule
No & Yes & $\mathbf{12}$ & $\mathit{0}$ & 0.91 & 0.827 & 0.725 & 0.572 & 0.186 & 0.0507 \\
No & Yes & $\mathbf{12}$ & $\mathit{4}$ & \textit{0.917} & \textit{0.84} & \textit{0.744} & \textit{0.597} & \textit{0.178} & \textit{0.045} \\
No & Yes & $\mathbf{12}$ & $\mathit{8}$ & 0.908 & 0.821 & 0.717 & 0.568 & 0.195 & 0.0523 \\
\midrule
No & Yes & $\mathbf{24}$ & $\mathit{0}$ & 0.928 & 0.862 & 0.777 & 0.634 & 0.154 & 0.04 \\
No & Yes & $\mathbf{24}$ & $\mathit{4}$ & 0.929 & 0.865 & 0.784 & 0.649 & 0.149 & 0.0395 \\
No & Yes & $\mathbf{24}$ & $\mathit{8}$ & \textit{0.932} & \textit{0.872} & \textit{0.795} & \textit{0.663} & \textit{0.144} & \textit{0.0349} \\
No & Yes & $\mathbf{24}$ & $\mathit{12}$ & 0.927 & 0.862 & 0.779 & 0.645 & 0.155 & 0.0414 \\
No & Yes & $\mathbf{24}$ & $\mathit{16}$ & 0.927 & 0.859 & 0.774 & 0.637 & 0.154 & 0.0385 \\
No & Yes & $\mathbf{24}$ & $\mathit{20}$ & 0.928 & 0.86 & 0.773 & 0.632 & 0.154 & 0.0392 \\
\midrule
No & Yes & $\mathbf{36}$ & $\mathit{0}$ & 0.934 & 0.876 & 0.799 & 0.664 & 0.141 & 0.0358 \\
No & Yes & $\mathbf{36}$ & $\mathit{4}$ & 0.935 & 0.878 & 0.806 & 0.677 & 0.138 & 0.0335 \\
No & Yes & $\mathbf{36}$ & $\mathit{8}$ & 0.934 & 0.879 & 0.807 & 0.681 & 0.139 & 0.0351 \\
No & Yes & $\mathbf{36}$ & $\mathit{12}$ & \textit{0.936} & \textit{0.882} & \textit{0.812} & \textbf{0.69} & \textit{0.134} & \textbf{0.0314} \\
No & Yes & $\mathbf{36}$ & $\mathit{16}$ & 0.935 & 0.879 & 0.808 & 0.682 & 0.136 & 0.0324 \\
No & Yes & $\mathbf{36}$ & $\mathit{20}$ & 0.934 & 0.876 & 0.802 & 0.676 & 0.138 & 0.0337 \\
No & Yes & $\mathbf{36}$ & $\mathit{24}$ & 0.934 & 0.875 & 0.8 & 0.671 & 0.139 & 0.0337 \\
No & Yes & $\mathbf{36}$ & $\mathit{28}$ & 0.934 & 0.875 & 0.8 & 0.672 & 0.14 & 0.0338 \\
No & Yes & $\mathbf{36}$ & $\mathit{32}$ & 0.934 & 0.873 & 0.796 & 0.665 & 0.141 & 0.0346 \\
\midrule
$\mathsf{pos}$ & $S\rightarrow R$ & $\mathbf{26.1}$ & $\mathit{0}$ & 0.926 & 0.86 & 0.775 & 0.634 & 0.157 & 0.0407 \\
$\mathsf{pos+neg}$ & $S\rightarrow R$ & $\mathbf{29}$ & $\mathit{10.4}$ & 0.931 & 0.869 & 0.79 & 0.658 & 0.147 & 0.0363 \\
$\mathsf{pos}$ & $R\rightarrow S$ & $\mathbf{33.6}$ & $\mathit{0}$ & 0.934 & 0.875 & 0.797 & 0.661 & 0.14 & 0.0366 \\
$\mathsf{pos+neg}$ & $R\rightarrow S$ & $\mathbf{35}$ & $\mathit{14.7}$ & \textbf{0.942} & \textbf{0.887} & \textbf{0.815} & 0.689 & \textbf{0.125} & 0.0319 \\
\midrule
No (Vavilala~\citeyear{Vavilala_2023_ICCV}) & Yes & \textbf{13.9} & \textit{0} & $0.869$ & $0.725$ & $0.565$ & $0.382$ & $0.266$ & $0.101$\\
No (Kluger~\citeyear{kluger2021cuboids}) & N/A & \textbf{-} & \textit{0} & $0.772$ & $0.627$ & $0.491$ & $0.343$ & $0.208$ & -\\
\bottomrule
\end{tabular}
}
\caption{\textbf{Baseline comparisons:} Ensembling strongly outperforms two recent SOTA methods, using the metrics reported by~\citet{kluger2021cuboids}, and using negative primitives in the ensemble produces further improvements. We show results with only positive primitives present $\mathsf{pos}$, three networks, $K^{total} \in \{\mathbf{12}, \mathbf{24}, \mathbf{36}\}$, as well as with positive and negative primitives $\mathsf{pos+neg}$, 18 networks, $K^- \in \{\mathit{0}, \mathit{4}, ..., \mathit{K}^{total}-\mathit{4}\}$. Our ensembles significantly outperform existing work. Further, we present results on the 18 methods we trained, where $K^{total}/K^-$ is shown. Even without ensembling, any individual method we trained performs better than the baselines. Notice that negative primitives are helpful on average.}
\label{tab:auc_NYU}
\end{table*}

\begin{figure*}[ht]
    \centering
    \begin{subfigure}[t]{0.48\linewidth}
        \centering
        \includegraphics[page=1, width=\linewidth]{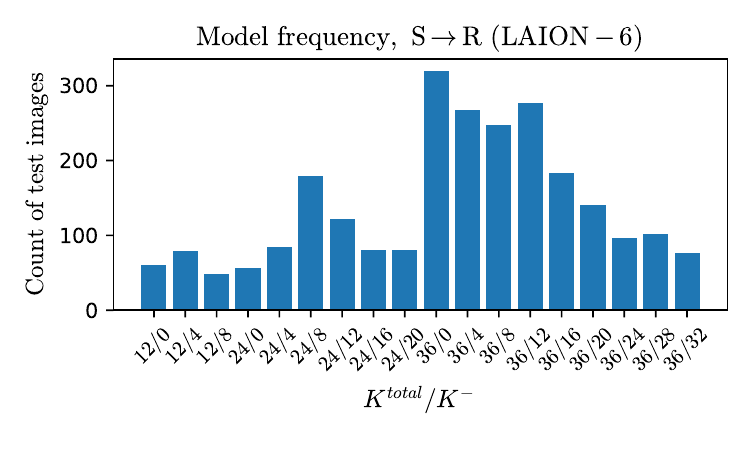} %
        \caption{Select then refine ensembling on LAION 6 faces.}
        \label{fig:laion6_hist_a}
    \end{subfigure}
    \hfill
    \begin{subfigure}[t]{0.48\linewidth}
        \centering
        \includegraphics[page=1, width=\linewidth]{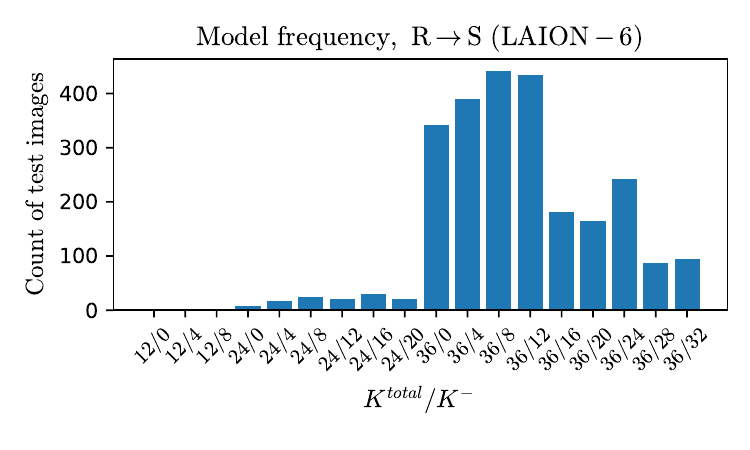} %
        \caption{Refine then select ensembling on LAION 6 faces}
        \label{fig:laion6_hist_b}
    \end{subfigure}
    \caption{Distribution of models chosen on LAION (6 faces), 2500 image test set. Models with negative primitives are often chosen, especially after finetuning.}
    \label{fig:laion6_hist}
\end{figure*}

\begin{figure*}[ht]
    \centering
    \begin{subfigure}[t]{0.48\linewidth}
        \centering
        \includegraphics[page=1, width=\linewidth]{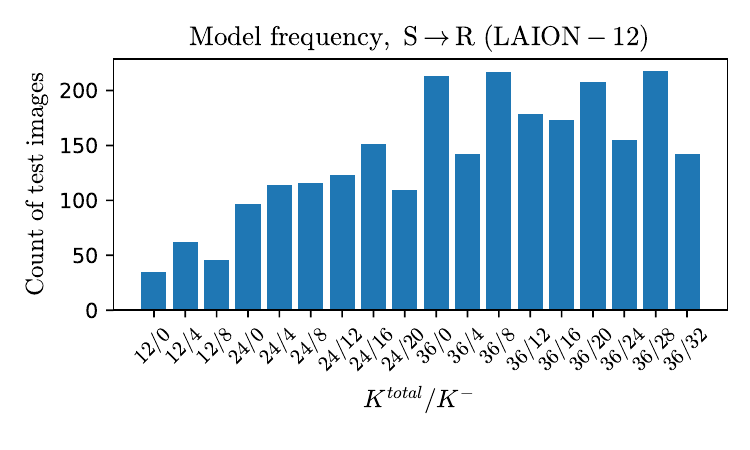} %
        \caption{Select then refine ensembling on LAION 12 faces.}
        \label{fig:laion12_hist_a}
    \end{subfigure}
    \hfill
    \begin{subfigure}[t]{0.48\linewidth}
        \centering
        \includegraphics[page=1, width=\linewidth]{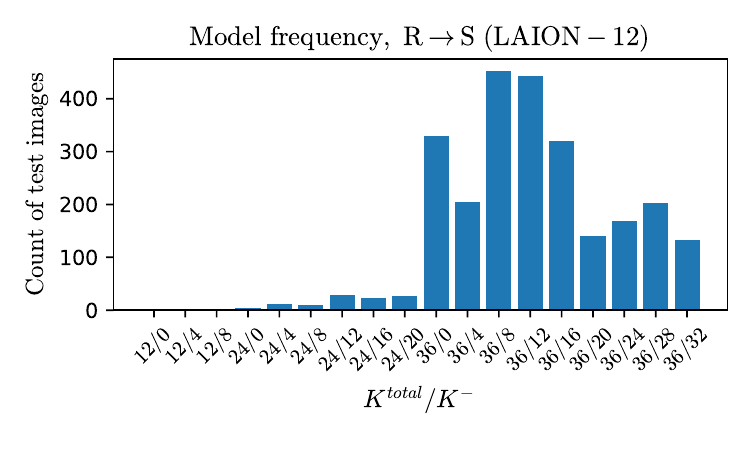} %
        \caption{Refine then select ensembling on LAION 12 faces}
        \label{fig:laion12_hist_b}
    \end{subfigure}
    \caption{Distribution of models chosen on LAION (12 faces), 2500 image test set. Models with negative primitives are often chosen, especially after finetuning.   }
    \label{fig:laion12_hist}
\end{figure*}

\begin{table*}[ht]
\centering
\resizebox{\textwidth}{!}{%

\begin{tabular}{ccrrcccccc}
\toprule
Ensemble & Refine & $K^{total}$ & $K^-$ & $\text{AUC}_{@50}\uparrow$ & $\text{AUC}_{@20}\uparrow$ & $\text{AUC}_{@10}\uparrow$ & $\text{AUC}_{@5}\uparrow$ & $\text{mean}_{cm}\downarrow$ & $\text{median}_{cm}\downarrow$\\
\midrule
No & Yes & $\mathbf{12}$ & $\mathit{0}$ & 0.953 & 0.904 & 0.841 & 0.75 & 0.128 & 0.0364 \\
No & Yes & $\mathbf{12}$ & $\mathit{4}$ & 0.953 & 0.905 & \textit{0.844} & \textit{0.755} & 0.127 & \textit{0.0345} \\
No & Yes & $\mathbf{12}$ & $\mathit{8}$ & \textit{0.954} & \textit{0.905} & 0.842 & 0.75 & \textit{0.125} & 0.037 \\
\midrule
No & Yes & $\mathbf{24}$ & $\mathit{0}$ & 0.963 & 0.924 & 0.87 & 0.79 & 0.104 & 0.0288 \\
No & Yes & $\mathbf{24}$ & $\mathit{4}$ & 0.964 & 0.925 & 0.873 & 0.793 & 0.101 & 0.0276 \\
No & Yes & $\mathbf{24}$ & $\mathit{8}$ & 0.964 & \textit{0.926} & \textit{0.876} & \textit{0.798} & 0.101 & \textit{0.0267} \\
No & Yes & $\mathbf{24}$ & $\mathit{12}$ & 0.963 & 0.923 & 0.87 & 0.791 & 0.104 & 0.0302 \\
No & Yes & $\mathbf{24}$ & $\mathit{16}$ & \textit{0.964} & 0.923 & 0.87 & 0.788 & \textit{0.101} & 0.0287 \\
No & Yes & $\mathbf{24}$ & $\mathit{20}$ & 0.963 & 0.921 & 0.867 & 0.785 & 0.104 & 0.0293 \\
\midrule
No & Yes & $\mathbf{36}$ & $\mathit{0}$ & 0.967 & 0.932 & 0.883 & 0.807 & 0.0965 & 0.0257 \\
No & Yes & $\mathbf{36}$ & $\mathit{4}$ & 0.967 & 0.933 & 0.886 & 0.811 & 0.0974 & 0.0272 \\
No & Yes & $\mathbf{36}$ & $\mathit{8}$ & \textit{0.968} & \textit{0.934} & \textit{0.887} & \textit{0.813} & \textit{0.0921} & \textbf{0.024} \\
No & Yes & $\mathbf{36}$ & $\mathit{12}$ & 0.967 & 0.933 & 0.886 & 0.813 & 0.0929 & 0.0242 \\
No & Yes & $\mathbf{36}$ & $\mathit{16}$ & 0.967 & 0.931 & 0.882 & 0.806 & 0.0932 & 0.0256 \\
No & Yes & $\mathbf{36}$ & $\mathit{20}$ & 0.966 & 0.929 & 0.879 & 0.802 & 0.0956 & 0.0272 \\
No & Yes & $\mathbf{36}$ & $\mathit{24}$ & 0.967 & 0.93 & 0.879 & 0.802 & 0.0928 & 0.0263 \\
No & Yes & $\mathbf{36}$ & $\mathit{28}$ & 0.963 & 0.924 & 0.872 & 0.793 & 0.104 & 0.0365 \\
No & Yes & $\mathbf{36}$ & $\mathit{32}$ & 0.962 & 0.923 & 0.871 & 0.793 & 0.106 & 0.0357 \\
\midrule
$\mathsf{pos}$ & $S\rightarrow R$ & $\mathbf{30.3}$ & $\mathit{0}$ & 0.963 & 0.925 & 0.872 & 0.794 & 0.104 & 0.0286 \\
$\mathsf{pos+neg}$ & $S\rightarrow R$ & $\mathbf{31.3}$ & $\mathit{10.6}$ & 0.965 & 0.927 & 0.877 & 0.8 & 0.0985 & 0.0273 \\
$\mathsf{pos}$ & $R\rightarrow S$ & $\mathbf{34.4}$ & $\mathit{0}$ & 0.967 & 0.932 & 0.882 & 0.805 & 0.0949 & 0.0262 \\
$\mathsf{pos+neg}$ & $R\rightarrow S$ & $\mathbf{35.4}$ & $\mathit{11.7}$ & \textbf{0.971} & \textbf{0.937} & \textbf{0.889} & \textbf{0.814} & \textbf{0.0827} & 0.0244 \\
\bottomrule
\end{tabular}

}
\caption{\textbf{Quantitative evaluation on LAION 6 face polytopes:} We train and ensemble models on a subset of LAION, with approx. 1.8M images in the training set and 2500 in the test set. We report error metrics defined in by~\citet{kluger2021cuboids}. Negative primitives remain useful, noting the italiciced error metrics in each block of $K^{total}$ always has negative primitives. Ensembling produces further improvements similar to NYUv2. Our method scales very well to in-the-wild scenes, producing even better metrics than NYUv2 given the larger dataset.}
\label{tab:auc_laion_6}
\end{table*}

\begin{table*}[ht]
\centering
\resizebox{\textwidth}{!}{%
\begin{tabular}{ccrrcccccc}
\toprule
Ensemble & Refine & $K^{total}$ & $K^-$ & $\text{AUC}_{@50}\uparrow$ & $\text{AUC}_{@20}\uparrow$ & $\text{AUC}_{@10}\uparrow$ & $\text{AUC}_{@5}\uparrow$ & $\text{mean}_{cm}\downarrow$ & $\text{median}_{cm}\downarrow$\\
\midrule
No & Yes & $\mathbf{12}$ & $\mathit{0}$ & 0.959 & 0.913 & 0.854 & 0.765 & 0.114 & 0.0339 \\
No & Yes & $\mathbf{12}$ & $\mathit{4}$ & 0.96 & 0.918 & \textit{0.863} & \textit{0.779} & 0.108 & \textit{0.0299} \\
No & Yes & $\mathbf{12}$ & $\mathit{8}$ & \textit{0.961} & \textit{0.918} & 0.862 & 0.777 & \textit{0.108} & 0.0301 \\
\midrule
No & Yes & $\mathbf{24}$ & $\mathit{0}$ & 0.967 & 0.931 & 0.881 & 0.804 & 0.0959 & 0.0268 \\
No & Yes & $\mathbf{24}$ & $\mathit{4}$ & 0.968 & 0.934 & 0.885 & 0.81 & 0.0912 & 0.0248 \\
No & Yes & $\mathbf{24}$ & $\mathit{8}$ & 0.968 & 0.933 & 0.885 & 0.811 & 0.0916 & 0.0246 \\
No & Yes & $\mathbf{24}$ & $\mathit{12}$ & \textit{0.969} & \textit{0.935} & \textit{0.887} & \textit{0.812} & \textit{0.0881} & 0.0243 \\
No & Yes & $\mathbf{24}$ & $\mathit{16}$ & 0.968 & 0.934 & 0.886 & 0.812 & 0.0898 & \textit{0.0243} \\
No & Yes & $\mathbf{24}$ & $\mathit{20}$ & 0.967 & 0.928 & 0.877 & 0.801 & 0.0927 & 0.0274 \\
\midrule
No & Yes & $\mathbf{36}$ & $\mathit{0}$ & 0.97 & 0.939 & 0.893 & 0.821 & 0.0872 & 0.0236 \\
No & Yes & $\mathbf{36}$ & $\mathit{4}$ & 0.971 & 0.94 & 0.895 & 0.823 & 0.0841 & 0.0227 \\
No & Yes & $\mathbf{36}$ & $\mathit{8}$ & 0.971 & 0.941 & 0.897 & 0.827 & 0.0829 & 0.0218 \\
No & Yes & $\mathbf{36}$ & $\mathit{12}$ & \textit{0.972} & \textit{0.942} & \textit{0.898} & \textit{0.828} & \textit{0.0816} & \textbf{0.0217} \\
No & Yes & $\mathbf{36}$ & $\mathit{16}$ & 0.971 & 0.94 & 0.896 & 0.825 & 0.0837 & 0.0221 \\
No & Yes & $\mathbf{36}$ & $\mathit{20}$ & 0.971 & 0.939 & 0.892 & 0.82 & 0.0845 & 0.0231 \\
No & Yes & $\mathbf{36}$ & $\mathit{24}$ & 0.971 & 0.939 & 0.894 & 0.822 & 0.0836 & 0.0227 \\
No & Yes & $\mathbf{36}$ & $\mathit{28}$ & 0.971 & 0.939 & 0.892 & 0.819 & 0.084 & 0.0232 \\
No & Yes & $\mathbf{36}$ & $\mathit{32}$ & 0.971 & 0.938 & 0.891 & 0.818 & 0.0851 & 0.0234 \\
\midrule
$\mathsf{pos}$ & $S\rightarrow R$ & $\mathbf{29.8}$ & $\mathit{0}$ & 0.967 & 0.933 & 0.884 & 0.808 & 0.0943 & 0.0262 \\
$\mathsf{pos+neg}$ & $S\rightarrow R$ & $\mathbf{31.2}$ & $\mathit{13.5}$ & 0.969 & 0.935 & 0.888 & 0.815 & 0.0884 & 0.0245 \\
$\mathsf{pos}$ & $R\rightarrow S$ & $\mathbf{35.3}$ & $\mathit{0}$ & 0.97 & 0.938 & 0.892 & 0.82 & 0.0867 & 0.0241 \\
$\mathsf{pos+neg}$ & $R\rightarrow S$ & $\mathbf{35.5}$ & $\mathit{13.2}$ & \textbf{0.974} & \textbf{0.944} & \textbf{0.899} & \textbf{0.829} & \textbf{0.0759} & 0.022 \\
\bottomrule
\end{tabular}

}
\caption{\textbf{Quantitative evaluation on LAION 12 face polytopes:} Most recent literature on primitive-fitting focuses on cuboids or parallelepipeds, but our model is capable of fitting polytopes of variable face count. All error metrics get better with more faces, which indicates more complex primitives yield more accurate representations. }
\label{tab:auc_laion_12}
\end{table*}

\begin{figure*}	[h]
    \centering
    \includegraphics[width=\linewidth]{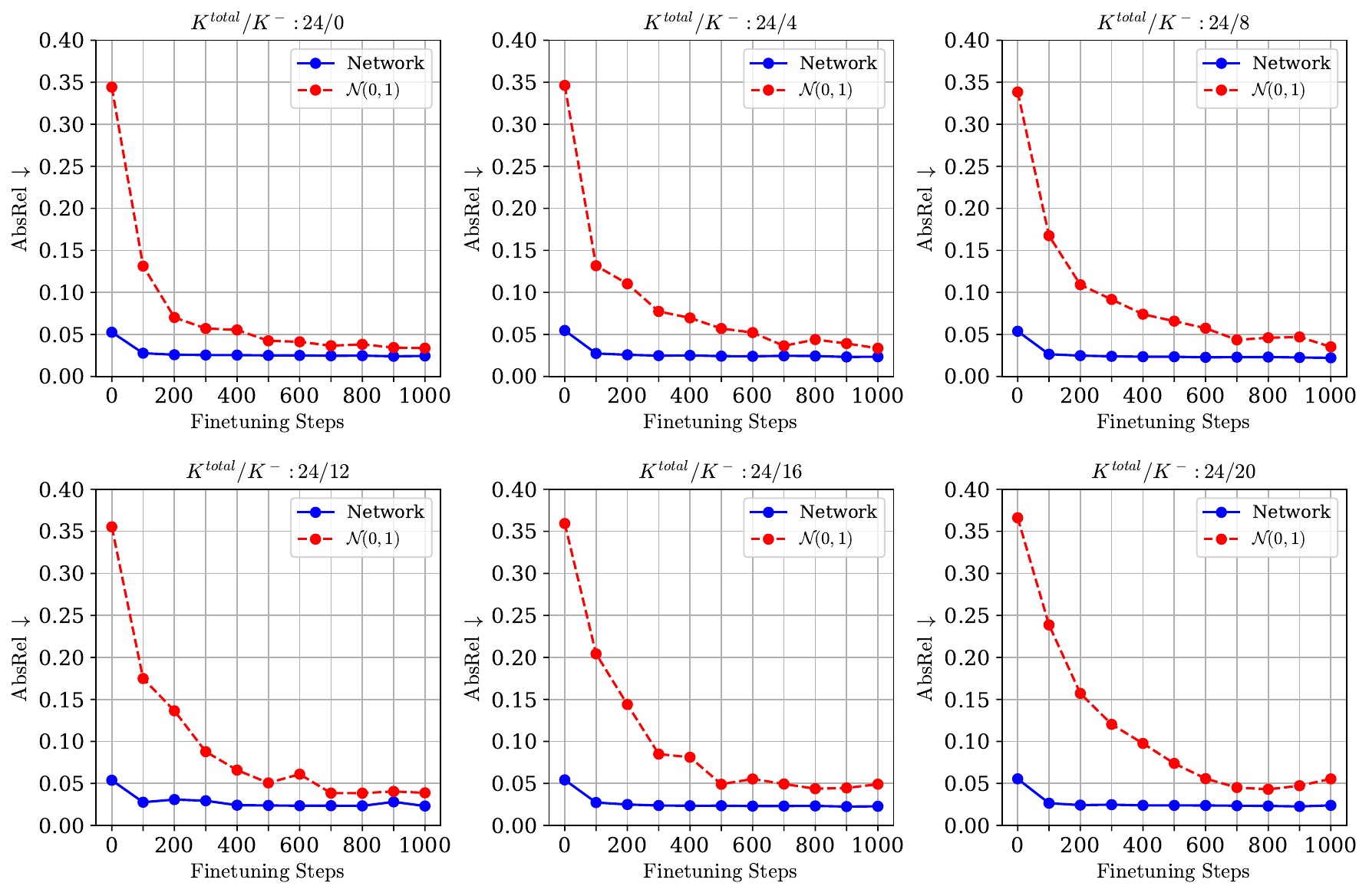}%
    \caption{Additional examples on the value of network start, on 100 LAION test images, $K^{total}=\mathbf{24}$.}
    \label{fig:FineTune24}
\end{figure*}  

\begin{figure*}	[h]
    \centering
    \includegraphics[width=\linewidth]{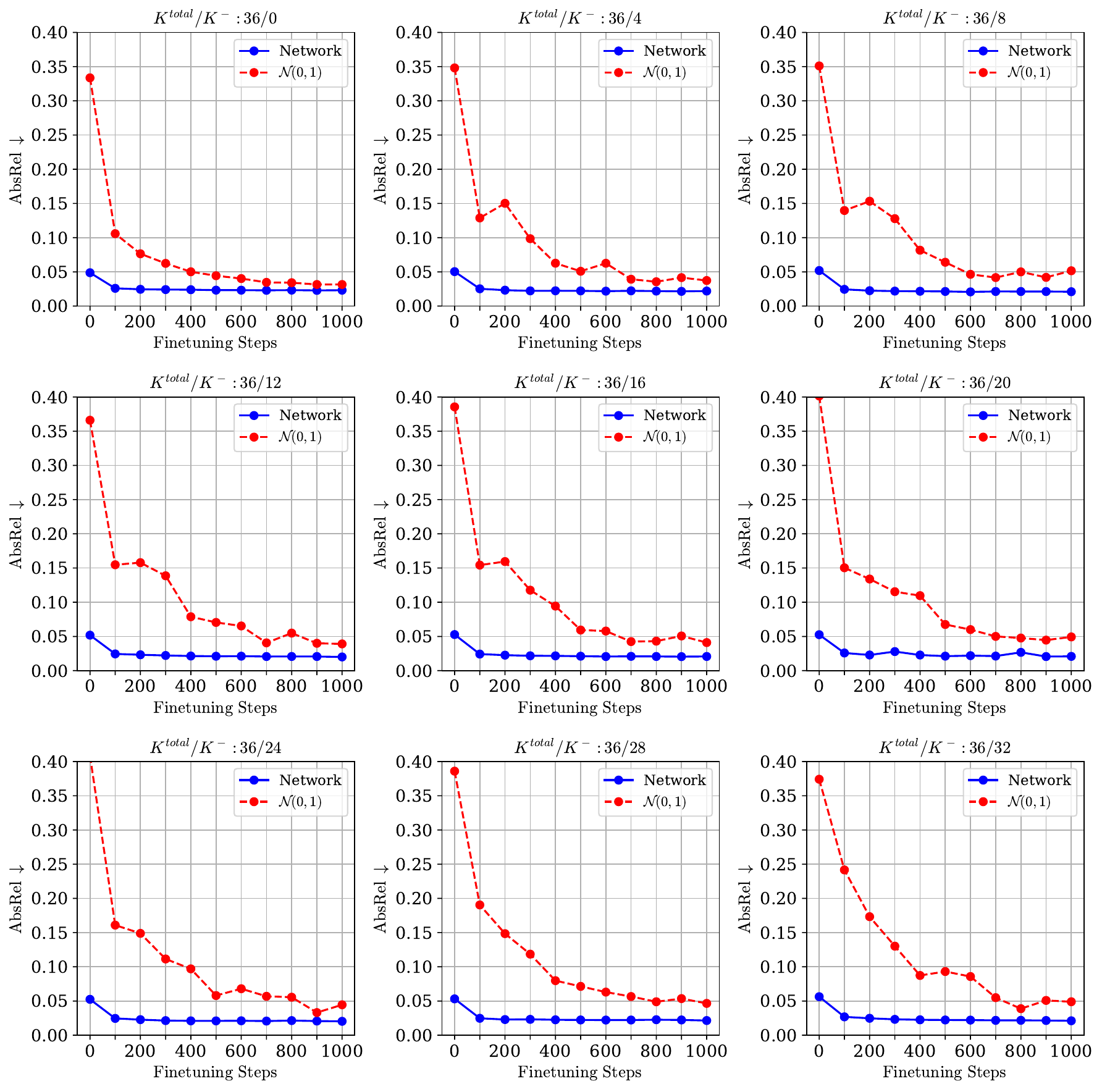}%
    \caption{Additional examples on the value of network start, on 100 LAION test images, $K^{total}=\mathbf{36}$.}
    \label{fig:FineTune36}
\end{figure*}

\begin{table*}[htb]
\centering
\resizebox{\textwidth}{!}{%
 \begin{tabular}{l c c c c c c}
\toprule
Ensemble & $K^{total}$ & $K^-$ & AbsRel$\downarrow$ & Normals Mean$\downarrow$ &Normals Median$\downarrow$ & SegAcc$\uparrow$ \\
\midrule
No & $\mathbf{12}$ & $\mathit{0}$ & 0.0622 & \textit{34.4} & 26.3 & 0.651 \\
No & $\mathbf{12}$ & $\mathit{4}$ & \textit{0.0597} & 34.5 & \textit{26.3} & \textit{0.68} \\
No & $\mathbf{12}$ & $\mathit{8}$ & 0.064 & 35.7 & 27.2 & 0.666 \\
\midrule
No & $\mathbf{24}$ & $\mathit{0}$ & 0.052 & 33 & 25 & 0.7 \\
No & $\mathbf{24}$ & $\mathit{4}$ & 0.0504 & 33 & 24.9 & 0.726 \\
No & $\mathbf{24}$ & $\mathit{8}$ & \textit{0.0486} & \textit{32.7} & \textit{24.6} & \textit{0.742} \\
No & $\mathbf{24}$ & $\mathit{12}$ & 0.0514 & 33.4 & 25.3 & 0.724 \\
No & $\mathbf{24}$ & $\mathit{16}$ & 0.0506 & 33.8 & 25.7 & 0.724 \\
No & $\mathbf{24}$ & $\mathit{20}$ & 0.0508 & 33.9 & 25.6 & 0.709 \\
\midrule
No & $\mathbf{36}$ & $\mathit{0}$ & 0.0484 & 32.3 & 24.4 & 0.723 \\
No & $\mathbf{36}$ & $\mathit{4}$ & 0.0467 & 32.3 & 24.3 & 0.752 \\
No & $\mathbf{36}$ & $\mathit{8}$ & 0.0469 & 32.2 & 24.2 & 0.757 \\
No & $\mathbf{36}$ & $\mathit{12}$ & \textit{0.0452} & \textit{32.1} & \textit{24} & \textbf{0.765} \\
No & $\mathbf{36}$ & $\mathit{16}$ & 0.0462 & 32.3 & 24.3 & 0.756 \\
No & $\mathbf{36}$ & $\mathit{20}$ & 0.0463 & 32.6 & 24.5 & 0.756 \\
No & $\mathbf{36}$ & $\mathit{24}$ & 0.0464 & 32.8 & 24.6 & 0.748 \\
No & $\mathbf{36}$ & $\mathit{28}$ & 0.0466 & 32.8 & 24.7 & 0.745 \\
No & $\mathbf{36}$ & $\mathit{32}$ & 0.047 & 33 & 24.8 & 0.735 \\
\midrule
$\mathsf{pos}$ $S\rightarrow R$ & $\mathbf{26.1}$ & $\mathit{0}$ & 0.0527 & 33 & 24.9 & 0.7 \\
$\mathsf{pos+neg}$ $S\rightarrow R$ & $\mathbf{29}$ & $\mathit{10.4}$ & 0.0492 & 32.9 & 24.7 & 0.733 \\
$\mathsf{pos}$ $R\rightarrow S$ & $\mathbf{33.6}$ & $\mathit{0}$ & 0.0476 & 32.3 & 24.4 & 0.72 \\
$\mathsf{pos+neg}$ $R\rightarrow S$ & $\mathbf{35}$ & $\mathit{14.7}$ & \textbf{0.0417} & \textbf{31.9} & \textbf{24} & 0.756 \\
\midrule
Vavilala \& Forsyth~\cite{Vavilala_2023_ICCV} & $\mathbf{13.9}$ & $\mathit{0}$ & 0.098 & 37.4 & 32.4 & 0.618 \\
\bottomrule
\end{tabular}

}
\caption{\textbf{Detailed error metrics on NYUv2.}}
\label{tab:nyu_full}
\end{table*}

\begin{table*}
\centering
\resizebox{\textwidth}{!}{%
\begin{tabular}{l c c c c c c}
\toprule
Ensemble & $K^{total}$ & $K^-$ & AbsRel$\downarrow$ & Normals Mean$\downarrow$ &Normals Median$\downarrow$ & $Neg\_per\_pos$ \\
\midrule
No & $\mathbf{12}$ & $\mathit{0}$ & 0.0297 & 35.6 & 29.3 & 0 \\
No & $\mathbf{12}$ & $\mathit{4}$ & 0.0295 & \textit{35.6} & \textit{29.1} & 1.42 \\
No & $\mathbf{12}$ & $\mathit{8}$ & \textit{0.029} & 35.8 & 29.2 & 4.13 \\
\midrule
No & $\mathbf{24}$ & $\mathit{0}$ & 0.0244 & 33.9 & 27.5 & 0 \\
No & $\mathbf{24}$ & $\mathit{4}$ & 0.0238 & 33.9 & 27.4 & 0.719 \\
No & $\mathbf{24}$ & $\mathit{8}$ & \textit{0.0235} & \textit{33.7} & \textit{27} & 1.54 \\
No & $\mathbf{24}$ & $\mathit{12}$ & 0.0242 & 34 & 27.4 & 2.39 \\
No & $\mathbf{24}$ & $\mathit{16}$ & 0.0235 & 34 & 27.4 & 3.94 \\
No & $\mathbf{24}$ & $\mathit{20}$ & 0.0242 & 34.2 & 27.5 & 8.04 \\
\midrule
No & $\mathbf{36}$ & $\mathit{0}$ & 0.0225 & 33 & 26.7 & 0 \\
No & $\mathbf{36}$ & $\mathit{4}$ & 0.0223 & 32.9 & 26.4 & 0.548 \\
No & $\mathbf{36}$ & $\mathit{8}$ & 0.0217 & \textit{32.9} & 26.3 & 1.01 \\
No & $\mathbf{36}$ & $\mathit{12}$ & 0.0218 & 32.9 & \textit{26.3} & 1.5 \\
No & $\mathbf{36}$ & $\mathit{16}$ & 0.0217 & 33.2 & 26.6 & 1.9 \\
No & $\mathbf{36}$ & $\mathit{20}$ & 0.0221 & 33.4 & 26.8 & 2.55 \\
No & $\mathbf{36}$ & $\mathit{24}$ & \textit{0.0215} & 33.3 & 26.7 & 3.81 \\
No & $\mathbf{36}$ & $\mathit{28}$ & 0.0236 & 33.8 & 27.1 & 5.49 \\
No & $\mathbf{36}$ & $\mathit{32}$ & 0.0243 & 33.9 & 27.3 & 10.8 \\
\midrule
$\mathsf{pos}$ $S\rightarrow R$ & $\mathbf{30.3}$ & $\mathit{0}$ & 0.0242 & 33.6 & 27.2 & 0 \\
$\mathsf{pos+neg}$ $S\rightarrow R$ & $\mathbf{31.3}$ & $\mathit{10.6}$ & 0.0229 & 33.4 & 26.8 & 2.04 \\
$\mathsf{pos}$ $R\rightarrow S$ & $\mathbf{34.4}$ & $\mathit{0}$ & 0.0221 & 33.1 & 26.7 & 0 \\
$\mathsf{pos+neg}$ $R\rightarrow S$ & $\mathbf{35.4}$ & $\mathit{11.7}$ & \textbf{0.0193} & \textbf{32.7} & \textbf{26.2} & 1.94 \\

\bottomrule
\end{tabular}

}
\caption{\textbf{Additional error metrics on LAION, 6 faces. The final column, $\mathrm{Neg\_per\_pos}$, evaluates the average number of negative primitives touching each positive primitive, quantitatively showing negative primitives active in the geometric abstraction.}}
\label{tab:laion_6_full}
\end{table*}

\begin{table*}
\centering
\resizebox{\textwidth}{!}{%
\begin{tabular}{l c c c c c c}
\toprule
Ensemble & $K^{total}$ & $K^-$ & AbsRel$\downarrow$ & Normals Mean$\downarrow$ &Normals Median$\downarrow$ & $Neg\_per\_pos$ \\
\midrule
No & $\mathbf{12}$ & $\mathit{0}$ & 0.0265 & 34.8 & 28.4 & 0 \\
No & $\mathbf{12}$ & $\mathit{4}$ & 0.0253 & \textit{34.3} & \textit{27.8} & 1.35 \\
No & $\mathbf{12}$ & $\mathit{8}$ & \textit{0.0252} & 34.4 & 27.8 & 3.89 \\
\midrule
No & $\mathbf{24}$ & $\mathit{0}$ & 0.0223 & 33.1 & 26.6 & 0 \\
No & $\mathbf{24}$ & $\mathit{4}$ & 0.0215 & 32.9 & 26.2 & 0.832 \\
No & $\mathbf{24}$ & $\mathit{8}$ & 0.0216 & 32.8 & 26.2 & 1.49 \\
No & $\mathbf{24}$ & $\mathit{12}$ & \textit{0.0208} & \textit{32.7} & \textit{26.1} & 2.48 \\
No & $\mathbf{24}$ & $\mathit{16}$ & 0.021 & 32.8 & 26.1 & 4.04 \\
No & $\mathbf{24}$ & $\mathit{20}$ & 0.0222 & 33.3 & 26.6 & 7.64 \\
\midrule
No & $\mathbf{36}$ & $\mathit{0}$ & 0.0203 & 32.2 & 25.7 & 0 \\
No & $\mathbf{36}$ & $\mathit{4}$ & 0.0199 & 32.2 & 25.6 & 0.648 \\
No & $\mathbf{36}$ & $\mathit{8}$ & 0.0196 & 32 & 25.3 & 1.1 \\
No & $\mathbf{36}$ & $\mathit{12}$ & \textit{0.0193} & \textit{32} & \textit{25.3} & 1.59 \\
No & $\mathbf{36}$ & $\mathit{16}$ & 0.0198 & 32.1 & 25.4 & 2.11 \\
No & $\mathbf{36}$ & $\mathit{20}$ & 0.0199 & 32.3 & 25.6 & 2.76 \\
No & $\mathbf{36}$ & $\mathit{24}$ & 0.0197 & 32.2 & 25.6 & 3.88 \\
No & $\mathbf{36}$ & $\mathit{28}$ & 0.0198 & 32.3 & 25.6 & 6.17 \\
No & $\mathbf{36}$ & $\mathit{32}$ & 0.0201 & 32.4 & 25.7 & 10.9 \\
\midrule
$\mathsf{pos}$ $S\rightarrow R$ & $\mathbf{29.8}$ & $\mathit{0}$ & 0.0218 & 32.7 & 26.3 & 0 \\
$\mathsf{pos+neg}$ $S\rightarrow R$ & $\mathbf{31.2}$ & $\mathit{13.5}$ & 0.0207 & 32.5 & 25.9 & 2.93 \\
$\mathsf{pos}$ $R\rightarrow S$ & $\mathbf{35.3}$ & $\mathit{0}$ & 0.0201 & 32.2 & 25.8 & 0 \\
$\mathsf{pos+neg}$ $R\rightarrow S$ & $\mathbf{35.5}$ & $\mathit{13.2}$ & \textbf{0.0178} & \textbf{31.8} & \textbf{25.2} & 2.53 \\

\bottomrule
\end{tabular}

}
\caption{\textbf{Additional error metrics on LAION, 12 faces. The final column, $\mathrm{Neg\_per\_pos}$, evaluates the average number of negative primitives touching each positive primitive, quantitatively showing negative primitives active in the geometric abstraction.}}
\label{tab:laion_12_full}
\end{table*}

\clearpage
\onecolumn 
\twocolumn  

\section{Common Questions}

\textbf{Q1. How is the method different from CvxNet?} \\
\textbf{A1.} While our approach draws inspiration from CvxNet, the technical setting is fundamentally different and introduces non-trivial challenges. CvxNet fits convex primitives to clean, segmented meshes or point clouds, often with full object geometry available. In contrast, our method fits primitives directly to \emph{in-the-wild} RGB-D scenes, where:
\begin{itemize}
    \item Geometry is partial (due to occlusion and unknown backs of objects).
    \item Scene segmentation may not be available.
    \item Lighting, noise, and clutter introduce significant fitting ambiguity.
\end{itemize}
To address this, we:
\begin{enumerate}
    \item Integrate \textbf{set-differencing (negative primitives)}---a first in primitive-fitting for real RGB scenes---allowing representation of concavities and occluded voids.
    \item Design a \textbf{test-time ensembling pipeline} to select the optimal primitive count per scene using only geometric consistency metrics.
    \item Introduce \textbf{improved optimization, sampling, and hyperparameters} so that even our smallest models (12 primitives, no negatives) outperform prior work \cite{Vavilala_2023_ICCV,kluger2021cuboids} in accuracy, segmentation, and latency.
\end{enumerate}

\vspace{0.5em}
\textbf{Q2. Why use negative primitives and difference operations for real scenes? Aren’t convex-only shapes sufficient?} \\
\textbf{A2.} Positive-only convex shapes cannot represent important concave features without excessive oversegmentation. Negative primitives allow parsimonious modeling of real-world voids and cavities (e.g., under desks, chair legs, hollow shelves) without increasing primitive count. Quantitatively:
\begin{itemize}
    \item \textbf{Segmentation Accuracy} improves from $0.723$ (positive-only) to $0.765$ (with negatives).
    \item Depth AbsRel improves significantly for the same primitive count when letting some primitives be negative (Table~\ref{tab:nyu_full}).
\end{itemize}
While some surfaces may appear ``carved,'' the improved geometry aligns better with ground-truth depth/normals. Moreover, the ensemble retains multiple reconstructions---users can choose purely convex outputs if visually preferred.

\vspace{0.5em}
\textbf{Q3. The ensembling strategy is a simple heuristic. Why is it a central contribution?} \\
\textbf{A3.} Fixed-count primitive-fitting fails to adapt to scene complexity variation. Our setting is unique because we can \textbf{quantitatively score reconstructions against input depth maps without ground truth primitives}---allowing a principled selection of primitive count at test time. This:
\begin{itemize}
    \item Consistently improves over the best single fixed-count model (Table~\ref{tab:Seg_table}).
    \item Produces results faster than prior SOTA~\cite{Vavilala_2023_ICCV}, even in the full ensemble mode (29.9s vs.\ 40s).
    \item Enables \emph{variable abstraction levels}, which prior works do not offer.
\end{itemize}
While ensembling is simple in concept, its \emph{problem-specific application} here solves a long-standing gap in scene decomposition---choosing the right model complexity without retraining.

\vspace{0.5em}
\textbf{Q4. Why convex polytopes instead of cuboids or superquadrics?} \\
\textbf{A4.} Convex polytopes offer:
\begin{itemize}
    \item \textbf{Flexibility}: By varying face count, they generalize cuboids/parallelepipeds while remaining convex.
    \item \textbf{Better optimization}: Represented as intersections of halfspaces, they are differentiable via logsumexp, unlike superquadrics, which often produce singularities, non-convexities, or self-intersections in certain parameter regimes.
    \item \textbf{Fair comparison}: Using the same primitive vocabulary as~\cite{Vavilala_2023_ICCV,kluger2021cuboids} ensures direct metric comparability.
\end{itemize}
Empirically, prior work~\cite{kluger2021cuboids} found cuboids outperform superquadrics in scene fitting; our parallelepipeds further improve accuracy (Table~\ref{tab:auc_NYU}).

\vspace{0.5em}
\textbf{Q5. The visual results sometimes appear over-fragmented or structurally implausible. How does this affect applications?} \\
\textbf{A5.} There is an inherent trade-off between \textbf{parsimonious, human-intuitive parses} and \textbf{high-fidelity geometric accuracy}.
\begin{itemize}
    \item For tasks like \emph{robotics grasp planning, simulation, or depth-to-image editing}, precise geometry may be preferred, even if primitives are more fragmented.
    \item For \emph{artist-friendly modeling}, fewer, larger convex shapes may be desired---our method supports this by training and exposing multiple abstraction levels (12--36 parts) and convex-only variants. Observe in Figs.~\ref{fig:qualNYUv2} and~\ref{fig:quallaion} how simple abstractions can be created with a few positive-only primitives, while complex (but more detailed) scene decompositions can be generated with a mixture of many primitives.
\end{itemize}
Prior works did not explore this trade-off or provide user-selectable abstraction.

\vspace{0.5em}
\textbf{Q6. How does the method compare in segmentation stability and runtime?} \\
\textbf{A6.} Stability: Despite increased geometric interactions from negative primitives, segmentation accuracy improves (SegAcc $0.756$ vs.\ $0.618$ in~\cite{Vavilala_2023_ICCV}). This suggests greater alignment with real object boundaries. \\
Runtime: Even the largest ensemble is \textbf{faster than prior SOTA}~\cite{Vavilala_2023_ICCV} and individual models run significantly faster (Table~\ref{tab:Seg_table}).

\vspace{0.5em}
\textbf{Q7. What about scalability and limitations?} \\
\textbf{A7.}
\begin{itemize}
    \item \textbf{Training scalability}: Current approach requires training separate models per primitive count; future work will explore unified variable-count architectures (e.g., transformer-based). It is non-trivial to avoid mode collapse with transformers because we do not know the GT positions of the primitives (unlike, say object detection). 
    \item \textbf{Primitive operations}: Only union and subtraction are explored; intersection is technically feasible but left for future work.
    \item \textbf{Failure cases}: Performance may degrade in scenes with extreme clutter, thin structures, or non-rigid shapes; examples are shown in Figures~\ref{fig:qualNYUv2} and~\ref{fig:quallaion} (worst ensemble members).
    \item \textbf{Resource constraints}: Ensembling increases inference cost, though individual models still surpass prior work in both speed and accuracy.
\end{itemize}

\end{document}